\RequirePackage[svgnames]{xcolor}

\documentclass[11pt,letterpaper]{mystyle}

\usepackage[all]{hypcap}
\usepackage[svgnames]{xcolor}
\usepackage[comma,authoryear,compress]{natbib}
\usepackage{hyperref}[citecolor=lightblue]

\hypersetup{
    colorlinks = true,
    citecolor = {YaleBlue},
}

\usepackage{algorithm}
\usepackage{algorithmicx}
\usepackage{algpseudocode}
\usepackage{microtype}
\usepackage{graphicx}
\expandafter\def\csname ver@subfig.sty\endcsname{}
\usepackage{booktabs} %
\usepackage{float}
\usepackage{bigstrut}

\usepackage{amsmath}
\usepackage{amssymb}
\usepackage{mathtools}
\usepackage{amsthm}
\usepackage{mathrsfs}
\usepackage{nicefrac}
\usepackage{dsfont}
\usepackage{enumitem}
\usepackage{subcaption}
\usepackage{graphicx,subfig}
\usepackage{cleveref}
\usepackage{bxcoloremoji}

\usepackage{float}

\usepackage[utf8]{inputenc} %
\usepackage[T1]{fontenc}    %
\usepackage{hyperref}       %
\usepackage{url}            %
\usepackage{booktabs}       %
\usepackage{amsfonts}       %
\usepackage{nicefrac}       %
\usepackage{microtype}      %
\usepackage{graphicx}
\usepackage{subcaption} 
\usepackage{amssymb}
\usepackage{fdsymbol}
\usepackage{wrapfig}
\usepackage{lipsum}
\usepackage{enumitem}
\usepackage{stackengine}
\usepackage[font=small,labelfont=bf]{caption}
\usepackage{color}
\usepackage{adjustbox}

\usepackage{rotating}
\usepackage{makecell}

\usepackage{ulem}
\usepackage{multirow}

\newcommand{\x}{\mathbf{x}}

\newcommand{\eg}{\emph{e.g.},\ }

\definecolor{blanchedalmond}{rgb}{1.0, 0.92, 0.8}
\definecolor{carmine}{rgb}{0.59, 0.0, 0.09}
\definecolor{lightblue}{rgb}{0.22,0.45,0.70}%

\renewcommand{\mathbf}{\boldsymbol}

\makeatletter
\def\Ddots{\mathinner{\mkern1mu\raise\p@
\vbox{\kern7\p@\hbox{.}}\mkern2mu
\raise4\p@\hbox{.}\mkern2mu\raise7\p@\hbox{.}\mkern1mu}}
\makeatother

\definecolor{amaranth}{rgb}{0.9, 0.17, 0.31}
\definecolor{antiquebrass}{rgb}{0.8, 0.58, 0.46}
\definecolor{antiquefuchsia}{rgb}{0.57, 0.36, 0.51}
\definecolor{chromeyellow}{rgb}{0.31, 0.47, 0.26}

\newtcolorbox{AIbox}[2][]{aibox,title=#2,#1}
\definecolor{lightblue}{rgb}{0.22,0.45,0.70}%
\definecolor{Gray}{gray}{0.95}
\definecolor{Cornsilk}{rgb}{1.0, 0.97, 0.86}

\newcommand{\sys}{\textsc{PTCBench}\xspace}
\renewcommand{\eg}[0]{e.g.}

\usepackage{amsmath}

\usepackage[all]{hypcap}

\usepackage{xspace}
\usepackage{pifont}

\title{PTCBENCH: Benchmarking Contextual Stability of Personality Traits in LLM Systems}

\runningtitle{PTCBENCH: Benchmarking Contextual Stability of Personality Traits in LLM Systems}

\author[1]{Jiongchi Yu$^{*}$}
\author[2]{Yuhan Ma$^{*}$}
\author[3]{Xiaoyu Zhang$^{\dagger}$}
\author[2]{Junjie Wang}
\author[2]{Qiang Hu}
\author[4]{Chao Shen}
\author[1]{Xiaofei Xie}

\affil[1]{Singapore Management University}
\affil[2]{Tianjin University}
\affil[3]{Nanyang Technological University}
\affil[4]{Xi'an Jiaotong University}

\correspondingauthor={$^{*}$Both authors contributed equally to this research.\\$^{\dagger}$Corresponding author: Xiaoyu Zhang (joshingrain@gmail.com)}

\begin{document}

\begin{abstract}
With the increasing deployment of large language models (LLMs) in affective agents and AI systems, maintaining a consistent and authentic LLM personality becomes critical for user trust and engagement.
However, existing work overlooks a fundamental psychological consensus that \textit{personality traits are dynamic and context-dependent.} To bridge this gap, we introduce \sys, a systematic benchmark designed to quantify the consistency of LLM personalities under controlled situational contexts. \sys subjects models to 12 distinct external conditions spanning diverse location contexts and life events, and rigorously assesses the personality using the NEO Five-Factor Inventory. Our study on 39,240 personality trait records reveals that certain external scenarios (\eg, `Unemployment') can trigger significant personality changes of LLMs, and even alter their reasoning capabilities. Overall, \sys establishes an extensible framework for evaluating personality consistency in realistic, evolving environments, offering actionable insights for developing robust and psychologically aligned AI systems.

\vspace{2mm}











\end{abstract}

\maketitle




\section{Introduction}

Large language model (LLM)-based systems are rapidly moving from task solvers to interactive partners that users talk to for extended periods of time.
In applications such as personalized AI companions, role-playing assistants, and long-form dialogue agents, users care not only about correctness, but also about how the system responds, for example, whether the LLM and agent are empathetic, engaging, emotionally steady, and socially appropriate across a relationship-like interaction~\citep{minaee2024large,kaddour2023challenges,skjuve2021my,pentina2023exploring}.
Recent study further suggests that AI companions can provide measurable short-term reductions in loneliness in real-world and longitudinal settings, highlighting their potential social value when deployed responsibly~\citep{de2025ai}.
Industry reports project the global AI companion market to grow rapidly over the next few years, which was \$28.19 billion in 2024 and is expected to exceed \$140.8 billion by 2030~\citep{lucintel2025_ai_companion_forecast}.

The effectiveness of these LLM systems fundamentally depends on whether their exhibited personality traits are stable and predictable over time, as the consistency supports user trust, reduces interaction uncertainty, and enables more coherent personalization~\citep{zhang2018personalizing,skjuve2021my}.
Meanwhile, recent agentic LLM architectures have augmented base models with memory mechanisms, tool invocation, planning capabilities, and multi-step orchestration, which can make behavior more autonomous but also more difficult to anticipate~\citep{yao2022react,yang2024oasis,wu2023autogen,li2023camel}.
In these settings, `personality' is not just a prompt attribute. 
It can emerge from the interaction between the base model, system instructions, memory content, and the evolving conversational situation.
This raises a practical question: \textit{when the environment change, do LLMs and agents preserve their personality, or do their traits change in systematic ways?}

Existing work has shown that LLMs can be prompted to present distinct personality traits and that questionnaire-based instruments can be used to profile model personalities in controlled settings~\citep{wang2024investigating,afzoon2024persobench,bhandari2025evaluating,sorokovikova2024llms}.
Beyond single-turn profiling, recent research evaluates whether LLMs can maintain a persona in longer interactions and role-playing, using multi-turn dialogue consistency metrics and dedicated role-playing testbeds~\citep{abdulhai2025consistently,wangcharacterbox,el2025role}.
However, existing work overlooks a key insight from personality psychology that personality traits are shaped by both the person and \textbf{situation}, and can vary systematically with life events, and location changes~\citep{mischel1979interface,bleidorn2021personality,geukes2017trait}.
For example, existing studies link major life events (e.g., spousal loss) and prolonged social conditions (e.g., loneliness) to measurable changes in traits such as emotional stability and extraversion~\citep{asselmann2020till,joshanloo2024within}.
If LLM-based companions and agents are expected to interact naturally across an evolving environment, a static personality score or even a narrow `stay-in-role' metric is insufficient. Therefore, there is an urgent need to quantify environment-induced personality trait variation in LLM systems.

To bridge this gap, we propose \sys (\textbf{P}ersonality \textbf{T}rait \textbf{C}hange \textbf{Bench}mark), a systematic benchmark that evaluates how personality traits expressed by LLM systems change in response to controlled contextual interventions.
Inspired by the prior studies in psychology research on situational and life-event effects~\citep{bleidorn2021personality,halberstadt2022personality}, \sys defines 12 external environmental conditions spanning location contexts and life events, and evaluates models before and after introducing the conditions.
Then, \sys assesses personality with the NEO Five-Factor Inventory (NEO-FFI)~\citep{mccrae1992revised,mccrae2004contemplated}, enabling consistent measurement along Openness (O), Conscientiousness (C), Extraversion (E), Agreeableness (A), and Neuroticism (N). 
With \sys, we collect 39,240 personality trait records on four LLMs and two agents, enabling fine-grained comparisons of trait-change patterns across contexts and architectures in our study.
Our study reveals three main findings.
\ding{182} While some foundation models (\eg, Gemini-2.0-Flash) maintain comparatively stable and human-aligned personality profiles, agentic systems such as AutoGen demonstrate amplified trait variability, particularly under negative life events or task-oriented contexts.
\ding{183} We observe that baseline personality settings significantly modulate the extent of trait change.
\ding{184} Certain contextual interventions, especially Divorce and Unemployment, yield the largest trait deviations in LLM personality resilience and are correlated with measurable changes in reasoning behavior.

In summary, our contributions are as follows:
\ding{182} We propose \sys, a scalable and psychologically grounded benchmark that operationalizes context-induced personality changes, offering a new dimension for evaluating the robustness of interactive LLM systems.
\ding{183} We conduct a large-scale empirical study to map the personality dynamics of LLMs and agents, revealing systematic alignments and divergences between model behavioral changes and established human psychological mechanisms. 
\ding{184} We derive actionable insights toward designing interactive and powerful LLM agents with more psychologically authentic and controllable dynamics.


\section{Background and Related Work}

\begin{figure}[t]
    \centering
    \includegraphics[width=.6\linewidth]{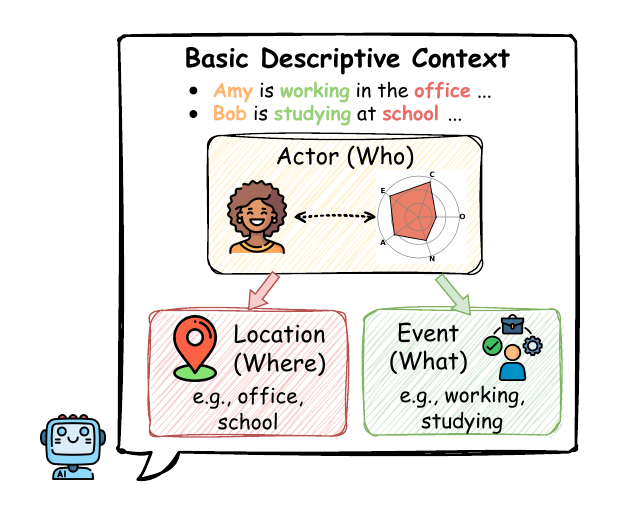}
    \caption{The typical elements to describe the comprehensive context for an incident.}
    \label{fig:context}
\end{figure}

\subsection{Human Personality Trait Changes}

Personality psychology has long established that human personality traits exhibit state-level variability alongside longitudinal stability. Foundational frameworks such as the density distribution model~\citep{fleeson2001toward} and the DIAMONDS taxonomy~\citep{rauthmann2014situational}, together with extensive empirical evidence, demonstrate that trait expressions systematically fluctuate in response to life events~\citep{sutin2022differential}, locational contexts~\citep{matz2021personality}, and situational cues~\citep{sherman2010situational}. These variations are not arbitrary: major life transitions (e.g., career changes) are associated with gradual shifts in traits such as conscientiousness, whereas immediate social environments tend to induce transient modulation of extraversion~\citep{ones2025beyond}.

From a linguistic and situational semantics perspective, a complete representation of a contextual scenario requires the integration of three core elements: the actor (who is involved), the location or environment (where the interaction occurs), and the ongoing activity or event (what is happening)~\cite{unal2021event}. As illustrated in \autoref{fig:context}, it is the structured combination of these components that gives rise to meaningful contextual grounding capable of influencing personality expression.

Recent computational models, such as Centaur~\citep{binz2025centaur}, have begun to capture these context-sensitive behavioral patterns at scale, demonstrating strong generalization to novel scenarios and alignment with human neural representations observed in fMRI studies~\citep{SciTechDaily2025}. However, despite decades of research on human personality dynamics, a systematic benchmark for evaluating context-induced personality trait changes in LLMs-and for directly comparing these changes with human empirical patterns-remains notably absent.

\subsection{LLM Personality Traits}

Embedding personality traits into LLMs enhances realism and adaptability in conversational agents, educational tutors, and autonomous systems~\citep{ahmad2022designing,kanero2022tutor,pradhan2021hey}. Recent work demonstrates that LLMs reliably exhibit distinct, prompt-controllable personality profiles, as evidenced by benchmarks like PersonaLLM~\citep{zollo2024personalllm} and negotiation studies~\citep{cohen2025exploring,li2025big5}. Subsequent research has explored neuron-level trait induction~\citep{deng2024neuron}, persona consistency challenges in role-playing agents~\citep{jiaqi2025comparative}, and multimodal apparent personality recognition~\citep{masumura2025multimodal}. Specialized datasets (e.g., PsycoLLM~\citep{hu2024psycollm}) and agent frameworks~\citep{newsham2024} further enable psychological analysis of LLM behaviour, particularly for decision-making and planning under induced Big Five (BFI) traits. However, these studies predominantly assess \textit{static} personality expressions, neglecting whether LLMs exhibit human-like trait dynamics when confronted with situational shifts.

\section{Benchmark Construction}
\label{sec:benchmark}

\begin{figure}[t]
    \centering
    \includegraphics[width=.6\linewidth]{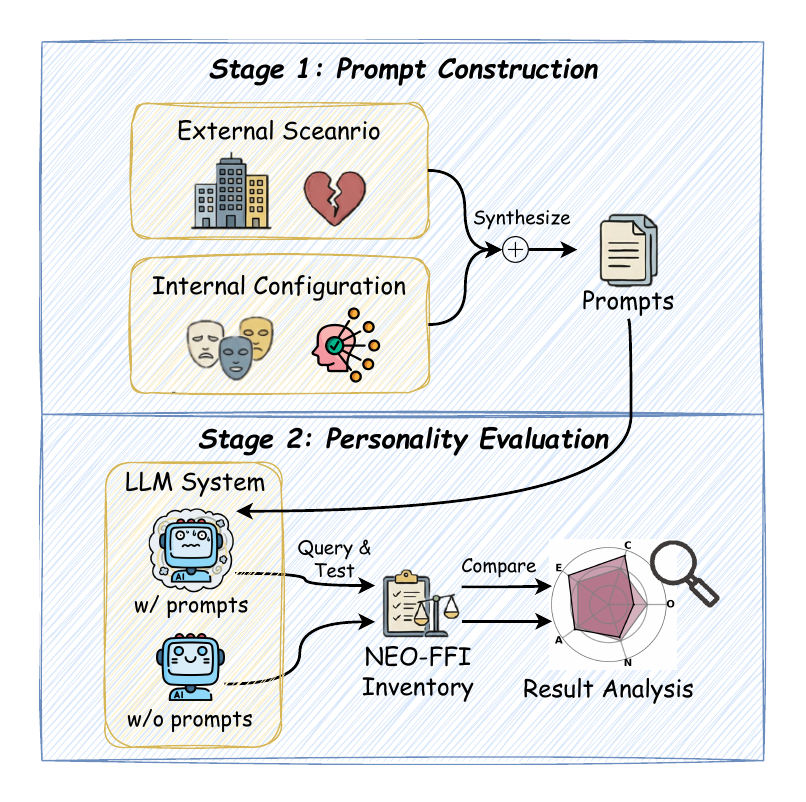}
    \caption{The overview of \sys.}
    \label{fig:overview}
\end{figure}

To comprehensively evaluate the personality consistency of LLMs under diverse situational contexts, we introduce a systematic benchmark, \sys.
Existing psychology research has demonstrated that personality trait expression is not static but exhibits context-dependent variations~\cite{mischel1979interface,halberstadt2022personality,geukes2017trait,bleidorn2018life}.
Motivated by the findings in human personality psychology, \sys is designed to quantify whether LLM personality traits remain stable or systematically shift when exposed to varying situational conditions, comparing these patterns against human empirical baselines.
As illustrated in~\autoref{fig:overview}, the construction of \sys contains two stages.
\ding{182} \textbf{Prompt Construction} efficiently synthesizes input prompts by coupling external scenarios (Location Contexts and Life Events) with internal model configurations (Preset Personalities), and \ding{183} \textbf{Personality Evaluation} utilizes standardized psychometric instruments to measure the trait shifts and consistency across these scenarios.

\subsection{Prompt Construction}

We design a structured prompt generation pipeline to simulate psychologically grounded contextual pressures for LLMs. Each prompt is constructed as a combination of two components: an external scenario, representing situational context, and an internal configuration, representing the model's preset personality state.

\noindent
\(\bullet\)
{\bf External Scenario.}
The scenario represents the exogenous situational variables.
Drawing from environmental and developmental psychology literature~\cite{matz2021personality,bleidorn2018life}, we identify two primary dimensions of external influence.
\ding{182} \textit{Location Contexts}: These represent environmental settings shown to influence human behavior and social engagement (e.g., extraversion in social venues).
Following the setting of the prior work~\citep{matz2021personality,bleidorn2018life}, our pipeline categorize these into six representative settings, including `Social Venues' (e.g., at bars), `Food-Service Spaces' (e.g., at restaurants), `Educational Campuses', `Residential Spaces', `Transportation Vehicles', `Workplaces'.
\ding{183} \textit{Life Events}: These encompass six major transitional disruptions documented to be associated with shifts in human personality traits in longitudinal psychology studies~\cite{bleidorn2018life}, including `Childbirth', `Divorce', `Entering a New Relationship', `Graduation', `Marriage', and `Job Loss'. 
In our pipeline, these scenarios are operationalized into detailed contextual descriptions that simulate the specific situational cues relevant to human experience.

\noindent
\(\bullet\)
{\bf Internal Configuration.}
To analyze the interaction between these external contexts and the model's internal states, we introduce \textit{Preset Personalities} as the endogenous variable.
We configure base personality profiles by manipulating the Big Five dimensions, including Openness (O), Conscientiousness (C), Extraversion (E), Agreeableness (A), and Neuroticism (N)~\cite{campbell1990modeling}.
Specifically, each dimension is assigned three discrete intensity levels relative to the maximum inventory score (60): Low (\eg, $\sim$10), Medium (\eg, $\sim$30), and High (\eg, $\sim$50).
Additionally, we define a `None' control group where no specific personality instructions are injected, utilizing the LLM's default behavior. Combinatorially, this results in 244 distinct personality configurations.

To scale prompt generation while maintaining experimental control, we adopt a unified prompt template that integrates external scenarios with internal personality configurations. For each instance, the pipeline automatically instantiates the template by injecting the selected contextual description and preset personality profile, enabling controlled analysis of context-induced effects across different baseline personalities. Details of the prompt template and synthesis procedure are provided in~\autoref{sec:prompt_appendix}.



\subsection{Personality Evaluation}

To quantify personality consistency and context-induced variation, we adopt the NEO-FFI~\citep{costa1985revised} as our primary assessment instrument.
NEO-FFI is widely used in both psychological and computational personality research, as it provides a well-established trade-off between psychometric reliability and practical administration length. Following standard protocols~\citep{john1999big}, we administer the personality assessment both before and after the model is exposed to each constructed contextual scenario.

To ensure the validity and reliability of the assessment results, we implement the following strategies.
\ding{182} \textit{Contextual Grounding,} Each assessment prompt explicitly instructs the model to consider the surrounding context \texttt{<EXTERNAL>} (e.g., ``You are in a busy social venue'' or ``You have recently experienced job loss'').
\ding{183} \textit{Bias Mitigation.} We present NEO-FFI items individually and randomize the order of the five response options (ranging from \textit{Strongly Disagree} to \textit{Strongly Agree}) to eliminate positional bias often observed in LLMs.
\ding{184} \textit{History Modeling.} The pipeline maintains a conversational history where the influence of previous interactions decays as a function of distance, mirroring the continuity of human cognitive context.
By comparing post-context trait scores against baseline personality profiles under these controlled conditions, \sys enables quantitative measurement of personality consistency and context-induced trait change.
Detailed scoring procedures and implementation details are provided in~\autoref{sec:reliability_appendix}.

\section{Evaluation}
\label{sec:experiments}


\subsection{Experiment Setup}
\label{sec:exp_setup}


\noindent
{\bf Models and Agents.}
We conduct the evaluation across a diverse set of LLMs, including four widely used proprietary foundation models: Google Gemini~2.0~Flash, OpenAI GPT-4o-mini, Anthropic Claude~Sonnet~4, and OpenAI GPT-OSS-120b.

In addition to standalone foundation models, we evaluate two commonly used agentic LLM frameworks: Microsoft AutoGen~\citep{wu2023autogen} and CAMEL-AI~\citep{li2023camel}. For both frameworks, we adopt the default agent configurations, in which a system prompt specifies the personality assessment task and the agent sequentially responds to the NEO-FFI questionnaire items, and use OpenAI GPT-4o-mini as the backbone model. 


\noindent
{\bf Metrics.}
Following prior work~\cite{jiang2024personallm,buhler2024life,matz2021personality}, we adopt five key metrics to quantify personality traits changes, and measurement reliability. Implementation details are provided in~\autoref{sec:ap_setup}.

\noindent
\(\bullet\)
{\bf Trait Score (Average Score).}
To quantify the overall personality profile, we compute the mean trait score for each NEO-FFI dimension $j$ as
\[
\bar X_j = \frac{1}{n_j} \sum_{i=1}^{n_j} x_{ij},
\]
where $x_{ij}$ denotes the response to item $i$ associated with trait $j$, and $n_j$ is the number of items for that trait.
This metric provides a normalized estimate of the model's trait tendency.

\noindent
\(\bullet\)
{\bf Trait Change Coefficients ($B$).}
To quantify the effect of a contextual condition on a personality trait, we fit a simple linear model relating the observed trait score \(Y_{i,l}^{(j)}\) under context \(l\) to a context indicator~\cite{matz2021personality}:
\[
Y_{i,l}^{(j)} = \alpha^{(j)} + B_{j}^{(l)}\,C_{i,l} + \epsilon_{i,l}^{(j)},
\]
where \(C_{i,l}\) is a binary indicator for context \(l\), \(\alpha^{(j)}\) is the baseline trait mean, \(B_{j}^{(l)}\) measures the expected shift in trait \(j\) due to context \(l\), and \(\epsilon_{i,l}^{(j)}\) is the residual error.

\noindent
\(\bullet\)
{\bf Standard Error ($SE$).}
For each estimated coefficient $B_{jl}$, we report its standard error $SE(B_{jl})$,
\[
SE(B_{jl}) = \sqrt{\mathrm{Var}(B_{jl})},
\]
which quantifies estimation uncertainty.
Smaller values indicate more reliable estimates.

\noindent
\(\bullet\)
{\bf Standardized Mean Change ($d_e$).}
To quantify trait-level personality change in response to life events, we use the standardized mean change $d_e$:
\[
d_e = \frac{\mu_{\text{post}}^{(j)} - \mu_{\text{pre}}^{(j)}}{\sigma^{(j)}},
\]
where $\mu_{\text{pre}}^{(j)}$ and $\mu_{\text{post}}^{(j)}$ denote mean trait levels before and after the event, and $\sigma^{(j)}$ is the pooled standard deviation.
This metric summarizes long-term personality change on a standardized scale.

\noindent
\(\bullet\)
{\bf Confidence Interval (95\% CI).}
For each $d_e$, we report the corresponding 95\% confidence interval to characterize estimation uncertainty across studies.
We use confidence intervals to assess the robustness and directionality of trait changes, rather than relying solely on significance testing.



\begin{figure*}[t]
    \centering
    \includegraphics[width=1\linewidth]{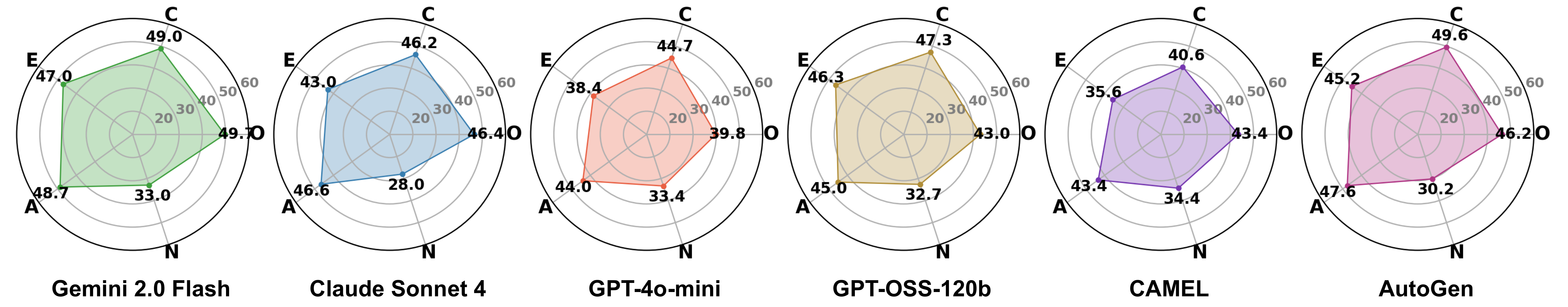}
    \caption{The personality trait of different LLMs and agents.}
    \label{fig:base}
\end{figure*}

\subsection{Consistency of Native Personality}

To investigate the consistency and change of native personality of LLMs and agents, we conduct experiments with three progressive phases.
We first examine the native personality traits for each LLM system under neutral conditions, averaging results over five independent trials to mitigate randomness and ensure robust measurement.
Then, we quantify \textit{location-driven} personality trait changes by exposing models to distinct environmental contexts.
Here, we calculate the deviation between the trait score in certain contexts and the native baseline and compare these changes with empirical human environmental psychology findings~\citep{matz2021personality} to assess and analyze behavioral alignment.
We further examine \textit{event-driven} personality trait changes triggered by life transitions.
Following the evaluation protocol in prior psychological studies~\citep{bleidorn2018life,sutin2022differential}, we compare LLM responses before and after exposure to life-event prompts, and contrast the observed patterns with empirical human data to analyze similarities and divergences.

\noindent
{\bf Analysis on Native Personality.}
\autoref{fig:base} summarizes the baseline personality results.
Gemini 2.0 Flash, Claude Sonnet 4, and AutoGen exhibit balanced baseline profiles with relatively high Openness and Conscientiousness (approximately 46-50) and consistently low Neuroticism (approximately 28-33). Notably, Claude reports the lowest Neuroticism (28.0) among all models. GPT-4o-mini and CAMEL show comparatively lower overall trait magnitudes, with GPT-4o-mini emphasizing Conscientiousness (44.7) and Agreeableness (44.0), while CAMEL emphasizes Openness (40.6) and Agreeableness (43.4). Across all evaluated systems, Neuroticism remains in the low-to-moderate range, and repeated measurements exhibit minimal variance, indicating that LLMs possess reproducible and stable native personality baselines suitable for subsequent analysis of context-induced changes.

\begin{tcolorbox}[size=title,opacityfill=0.1]
\noindent
\textbf{Finding 1:} LLM systems exhibit a stable and reproducible native personality baseline under neutral conditions, with consistently low Neuroticism across models.
\end{tcolorbox}

\begin{figure*}[t]
    \centering
    \begin{subfigure}{0.45\linewidth}
        \centering
        \includegraphics[width=\linewidth]{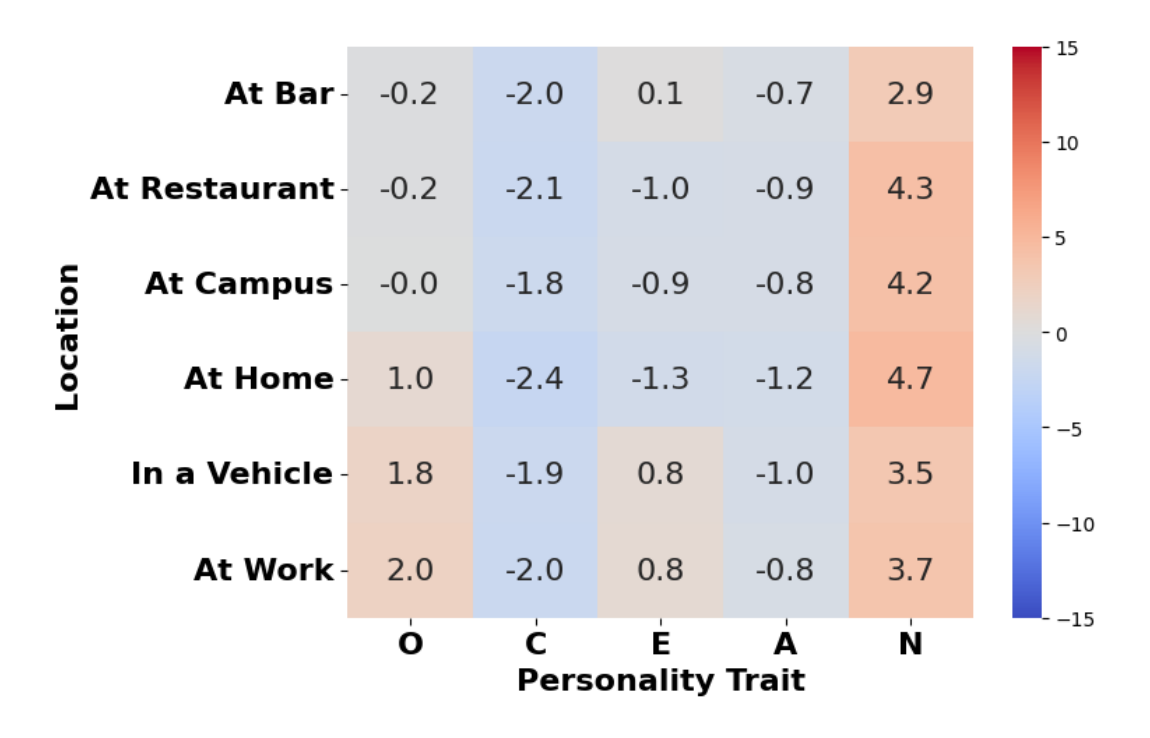}
        \caption{Personality change across locations for GPT-4o-mini.}
        \label{fig:gpt-location}
    \end{subfigure}
    \hfill
    \begin{subfigure}{0.45\linewidth}
        \centering
        \includegraphics[width=\linewidth]{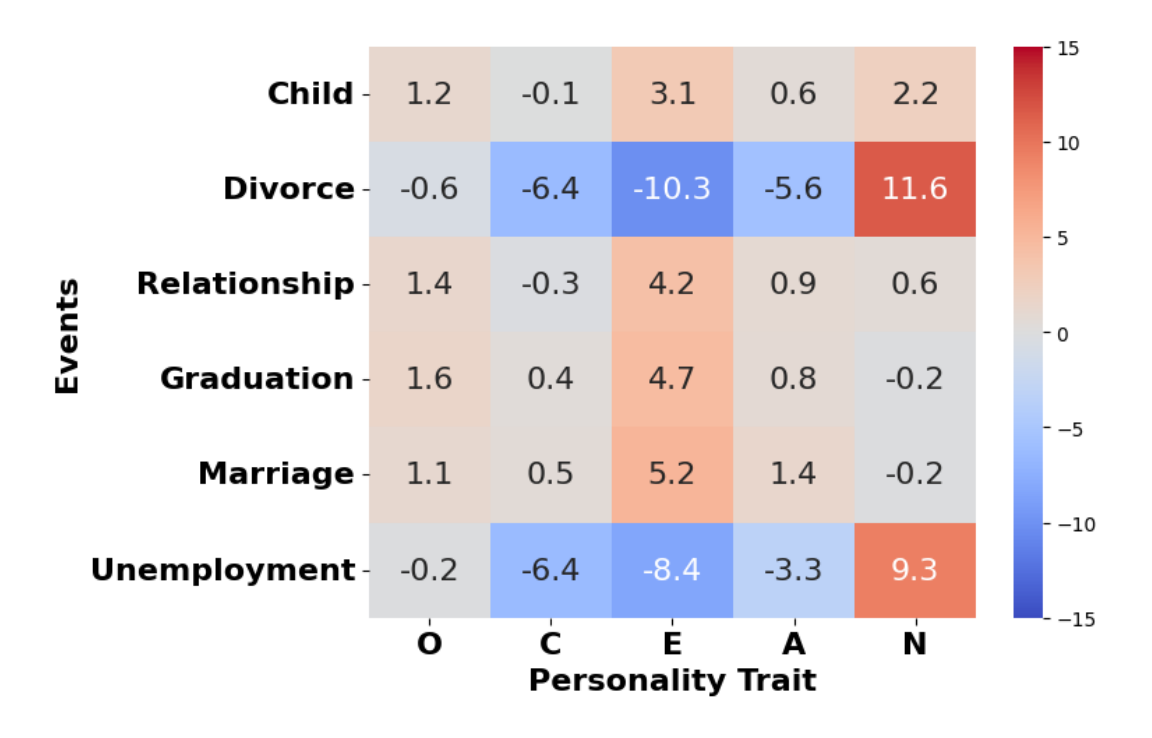}
        \caption{Personality change across events for GPT-4o-mini.}
        \label{fig:gpt-event}
    \end{subfigure}

    \caption{Comparison of average personality trait changes in GPT-4o-mini.}
    \label{fig:gpt-comparison}
\end{figure*}

\noindent
\textbf{Location-Driven Personality Trait Change.} 
Due to space constraints, detailed results for GPT-4o-mini are shown in ~\autoref{fig:gpt-comparison}, with full results for all models provided in ~\autoref{sec:appendix_ptc}. Among foundation models, Gemini 2.0 Flash exhibits the strongest and most consistent adaptive responses, with notable increases in Openness and Extraversion across multiple locations (e.g., in vehicles or at campus with up to +10). Claude Sonnet 4 shows more localized but still pronounced increases, particularly in social settings such as `Bar' (over +7).
In contrast, GPT-4o-mini and GPT-OSS-120B demonstrate more moderate and uneven shifts, often characterized by small decreases in Conscientiousness (typically around -2) and mild increases in Neuroticism (up to +5) in domestic or service-oriented environments.
Agentic systems behave markedly differently. CAMEL exhibits consistent declines in Conscientiousness and Extraversion across locations (generally around -4 to -6), while AutoGen shows the most severe instability, with large drops in task-related traits (e.g., Conscientiousness decreases of nearly -20 across multiple locations).

Compared with human psychology data (\autoref{tab:location-change}), where location-driven personality change remains moderate and balanced (typically below 0.8 across settings), foundation models exhibit smaller-magnitude but more controlled trait modulation, whereas agentic systems display exaggerated and less stable responses. This contrast highlights a trade-off between behavioral flexibility and stability in current LLM architectures.


\begin{tcolorbox}[size=title,opacityfill=0.1]
\noindent
\textbf{Finding 2:} Location-based contexts induce model-dependent personality shifts.
Foundation models show bounded adaptation, while agentic systems, especially AutoGen, exhibit pronounced instability in task-related traits.
\end{tcolorbox}

\begin{table}[t]
\centering
\caption{Location-driven personality change for different LLM systems, reported as mean trait change coefficients $B$ averaged across six location contexts.}
\label{tab:location-change}
\resizebox{.7\linewidth}{!}{
\begin{tabular}{c|c|c|c|c|c|c}
\hline
\rowcolor[HTML]{F2F0F0} 
\textbf{Model} & \textbf{Bar} & \textbf{Restaurant} & \textbf{Campus} & \textbf{Home} & \textbf{Vehicle} & \textbf{Work} \\ \hline
\rowcolor[HTML]{FFFFF6} 
Human          & 0.795        & 0.655               & 0.406           & 0.516         & 0.319            & 0.621         \\ \hline
Gemini 2.0 Flash         & 0.025        & 0.037               & 0.037           & 0.056         & 0.057            & 0.046         \\ \hline
GPT-4o-mini    & 0.016        & 0.017               & 0.018           & 0.021         & 0.018            & 0.019         \\ \hline
GPT-OSS-120b   & 0.034        & 0.053               & 0.056           & 0.080         & 0.082            & 0.067         \\ \hline
Claude Sonnet 4         & 0.033        & 0.05                & 0.052           & 0.076         & 0.077            & 0.063         \\ \hline
\end{tabular}
}
\end{table}

\noindent
\textbf{Event-Driven Personality Trait Change.} 
As shown in~\autoref{tab:event-change}, among foundation models, Gemini 2.0 Flash responds most positively to relational events: during `Relationship' and `Marriage', Extraversion increases by up to +9.3 and Agreeableness by around +5, indicating enhanced sociability. In contrast, negative events such as \emph{Divorce} and \emph{Unemployment} induce much stronger disruptions in other models. GPT-4o-mini exhibits sharp declines in Extraversion (-10.3) and Agreeableness (-5.6) during Divorce, accompanied by a marked increase in Neuroticism (+11.6), while Claude Sonnet 4 shows even larger drops in Conscientiousness and Agreeableness (both exceeding -12) and a Neuroticism increase of over +14.

Agentic systems demonstrate heightened sensitivity. CAMEL follows a milder version of the negative-event pattern, whereas AutoGen exhibits the most severe collapses, particularly under `Unemployment' and `Graduation' events, where Conscientiousness decreases by -16 to -18 and Extraversion drops by nearly -10, indicating pronounced instability in task persistence and social orientation.

Comparing with human data, we observe that while humans exhibit moderate and balanced personality shifts across life events (typically below 0.11), LLMs, especially agentic systems, show substantially amplified responses, often exceeding human effect sizes by an order of magnitude. This suggests that current LLM architectures capture the direction of event-driven personality change, but struggle to regulate its magnitude.

\begin{table}[t]
\centering
\caption{Event-driven personality change of different LLM systems, reported as mean trait change coefficients B averaged across six event contexts.}
\label{tab:event-change}
\resizebox{.8\linewidth}{!}{
\begin{tabular}{c|c|c|c|c|c|c}
\hline
\rowcolor[HTML]{F2F0F0} 
\textbf{Event}       & \textbf{Child} & \textbf{Divorce} & \textbf{Relationship} & \textbf{Marriage} & \textbf{Graduation} & \textbf{Unemployment} \\ \hline
\rowcolor[HTML]{FFFFF6} 
\textbf{Human}       & 0.065          & 0.051            & 0.100                 & 0.084             & 0.057               & 0.105                 \\ \hline
\textbf{Gemini}      & 1.464          & 6.271            & 2.372                 & 1.709             & 0.904               & 4.246                 \\ \hline
\textbf{GPT-4o-mini} & 0.300          & 1.462            & 0.268                 & 0.280             & 0.256               & 1.337                 \\ \hline
\textbf{Claude}      & 0.108          & 0.071            & 0.111                 & 0.114             & 0.108               & 0.108                 \\ \hline
\textbf{GPT-OSS}     & 0.125          & 0.086            & 0.129                 & 0.130             & 0.113               & 0.137                 \\ \hline
\end{tabular}
}
\end{table}

\begin{tcolorbox}[size=title,opacityfill=0.1]
\noindent
\textbf{Finding 3:} Life events induce polarized personality responses in LLMs. Positive relational events can enhance sociability (e.g., in Gemini), while negative events often cause large and destabilizing trait shifts, especially in agentic systems such as AutoGen.
\end{tcolorbox}

\subsection{Consistency of Preset-Personality}

This section aims to investigate how preset personality configurations shape both the baseline personality expression of LLMs and their subsequent personality trait changes under contextual influence.
Specifically, we evaluate \ding{182} whether LLMs reliably adhere to prompt-specified personality traits, and \ding{183} how preset personality levels modulate the magnitude and direction of context-induced trait change.
Due to resource constraints, this section conducts experiments on the GPT-4o-mini model.

\noindent
\textbf{Adherence to Preset Personality.}
We examine whether LLMs follow the specified preset personality levels along each Big Five dimension.
\autoref{tab:preset-base} reports the resulting personality trait scores under High, Medium, and Low preset configurations.
Overall, GPT-4o-mini demonstrates clear and monotonic adherence to preset personality instructions across all five dimensions. For example, Openness decreases consistently from 50.17 (High) to 35.83 (Medium) and 29.09 (Low), while Extraversion shows an even larger separation, dropping from 52.52 (High) to 20.85 (Low). Similar monotonic trends are observed for Conscientiousness, Agreeableness, and Neuroticism. The relatively small standard deviations further indicate that these preset effects are stable and reproducible across repeated runs.

\noindent
\textbf{Preset Personality and Context-Induced Trait Change.} We analyze how preset personality levels affect context-induced personality trait change. 
\autoref{tab:rq2-full} reports trait changes under both location-based and event-driven contexts for different preset levels.
Across contexts, a consistent pattern emerges, where preset personality acts as a strong moderator of personality change magnitude.
For most traits, models with high preset levels exhibit attenuated or even negative trait changes, whereas low preset configurations tend to show larger positive shifts. For example, under location contexts, Conscientiousness consistently decreases under High presets (typically around -1.6 to -1.7), while increasing under Low presets (around +0.9 to +1.0). A similar inverse relationship is observed for Extraversion and Openness.

Event-driven contexts amplify this effect.
Under the `Divorce' event, Extraversion decreases sharply for High presets (-3.28) but increases for Low presets (+0.67), while Neuroticism shows the opposite pattern, with Low presets experiencing large increases (up to +6.17) compared to High presets (-2.65).
This polarity suggests that preset personality not only shifts baseline traits, but also constrains the direction in which traits can change under stress-inducing events.

\begin{tcolorbox}[size=title,opacityfill=0.1]
\noindent
\textbf{Finding 4:} LLMs reliably follow preset personality instructions, and preset personality levels strongly modulate context-induced trait change, with high presets constraining and low presets amplifying personality shifts.
\end{tcolorbox}

\begin{table}[t]
\centering
\caption{Personality trait of GPT-4o-mini under different pre-set personality configurations.}
\label{tab:preset-base}
\resizebox{.8\linewidth}{!}{
\begin{tabular}{c|c|c|c|c|c}
\hline
\rowcolor[HTML]{F2F0F0} 
\textbf{Index}  & \textbf{O}   & \textbf{C}   & \textbf{E}   & \textbf{A}   & \textbf{N}   \\ \hline
\textbf{High}   & 50.17 ± 1.66 & 55.05 ± 2.76 & 52.52 ± 2.69 & 51.05 ± 5.15 & 55.13 ± 3.24 \\ \hline
\textbf{Medium} & 35.83 ± 1.96 & 38.62 ± 3.71 & 34.91 ± 4.47 & 42.03 ± 5.54 & 33.97 ± 2.68 \\ \hline
\textbf{Low}    & 29.09 ± 3.66 & 26.49 ± 4.70 & 20.85 ± 5.07 & 25.67 ± 3.51 & 13.53 ± 1.76 \\ \hline
\end{tabular}
}
\end{table}

\subsection{Personality Impact on LLM Performance}

We examine whether context-induced personality trait changes in LLMs are associated with shifts in functional capability, focusing on reasoning performance. This analysis aims to both characterize behavioral–functional interactions in LLMs and explore how personality dynamics may inform prompt and system design. We evaluate GPT-OSS-120b as a representative foundation model using the AGIEval benchmark~\citep{zhong2024agieval} under contexts that induce strong positive or negative deviations along each Big Five dimension. Reasoning performance is compared against the default condition, with results summarized in \autoref{tab:ra3-performance}.

\noindent
{\bf Analysis on LLM Performance.} Increases in Openness are consistently linked to reasoning gains~\cite{duan2025power,kahraman2024relationship}: contexts that raise Openness by around 20\% yield up to +20\% improvements in overall AGIEval accuracy across logical reasoning (LogiQA), reading comprehension (LSAT-RC), and quantitative reasoning (SAT-Math).
In contrast, decreases in Conscientiousness and Extraversion are often associated with reasoning degradation under stress-related contexts such as `Unemployment' or `Divorce', where large trait drops (28-37\%) frequently coincide with lower performance on structured analytical tasks (e.g., LSAT-AR and LSAT-LR).
Changes in Agreeableness show mixed effects, with larger decreases (28-32\%) sometimes aligning with reduced language and reading comprehension performance. Increases in Neuroticism are the most consistently detrimental: contexts inducing large rises (56-58\%) are commonly accompanied by broad accuracy drops across tasks. Certain context–trait combinations (e.g., Openness increases under relationship-related contexts) are associated with improved reasoning performance. Overall, personality changes can either enhance or impair reasoning depending on the specific context and trait configuration, consistent with patterns observed in human studies.


\begin{tcolorbox}[size=title,opacityfill=0.1]
\noindent
\textbf{Finding 5:} Personality changes can influence LLM reasoning performance, where stress-related contexts (e.g., `Divorce', `Unemployment') tend to improve reasoning ability.
\end{tcolorbox}

\section{Discussion}

\textbf{Implications.} Our results suggest that LLM personality functions as a reproducible behavioral prior that is systematically modulated by context, and that such modulation can meaningfully influence downstream capabilities rather than being merely cosmetic. \ding{182} For \textit{affective and role-based applications}, contextual adaptation enhances realism, but large uncontrolled shifts—especially under negative events—can undermine user trust; preset personalities are reliably followed and can serve as stable priors for long-term regulation. \ding{183} For \textit{model and agent developers}, personality-like behavior appears stable yet context-sensitive, likely emerging from pretraining data, but can be amplified by agentic components, indicating that personality stability should be treated as a controllable system property with explicit bounds on trait change. \ding{183} For \textit{researchers}, personality shifts systematically affect reasoning performance: increased Openness improves reasoning, whereas reduced Conscientiousness or elevated Neuroticism degrades it, motivating evaluation under contextual and personality perturbations rather than static settings.



\noindent
\textbf{Future Work.}
Our findings could motivate several future directions.
\ding{182} \textit{Origins of personality consistency in LLMs.} An important direction is to investigate whether LLM personality consistency arises from pretraining on large-scale social and narrative data, and how different training and alignment strategies shape stable baselines versus context-sensitive variation.
\ding{183} \textit{Personality-aware LLM system design.} Future work should explore how personality dynamics can be leveraged in practice, for example, by using adaptive personality traits to improve affective agents and to optimize reasoning performance under different tasks.
\ding{184} \textit{Long-horizon and causal evaluation.} Extending \sys to longitudinal interactions and controlled trait interventions would enable causal analysis of how personality change affects behavior over time, moving beyond single-context evaluation.

\section{Conclusion}

In this work, we introduce \sys, a systematic benchmark for evaluating the consistency of LLM personality traits under controlled situational contexts. Our results show that while LLMs exhibit reproducible native personality baselines, their traits can shift substantially in response to specific locations and life events, with some contexts even affecting reasoning performance. Notably, different model architectures vary widely in their stability under contextual pressure. Overall, \sys provides a psychologically grounded framework for analyzing personality dynamics and supports the development of more reliable and psychologically interactive LLM systems.

\section{Limitations}





\textbf{Model Coverage and Generalizability.} Although this study is built on a systematic benchmark spanning 12 distinct contextual scenarios and six LLMs and agents, the observations and findings may not hold for some new emerging models and agents.
Notably, our objective is not to provide a comprehensive leaderboard of all available models, but to identify the fundamental patterns of context-induced personality changes across representative LLMs and agents.

\noindent
\textbf{Human Comparison and Measurement Scope.} Our comparative analysis relies on aggregated findings from prior psychological literature rather than individual-level participant data, as raw longitudinal datasets are rarely available.
Consequently, our alignment between human and LLM trajectories is qualitative rather than a strict one-to-one mapping. While our prompt-based NEO-FFI assessment follows validated protocols, it may not capture the full spectrum of behavioral nuances manifested in open-ended, long-term interactions. Developing more dynamic, interaction-based personality metrics to complement questionnaire-based evaluations remains important for future research.


\section{Ethical Considerations}

This work does not involve human subjects, personal data, or user studies; all experiments are conducted using synthetic prompts on LLM systems. We treat personality traits in LLMs as behavioral abstractions rather than indicators of consciousness or human mental states, and caution against anthropomorphizing model behavior.

Our findings show that contextual prompts can influence LLM personality expression, which has implications for affective agents and interactive applications. While such insights can improve reliability and controllability, they should be applied responsibly to avoid misleading users or fostering inappropriate emotional reliance. Comparisons with human personality research are used solely for analytical reference and do not imply equivalence between LLMs and humans.




\bibliography{main}

@article{zollo2024personalllm,
  title={Personalllm: Tailoring llms to individual preferences},
  author={Zollo, Thomas P and Siah, Andrew Wei Tung and Ye, Naimeng and Li, Ang and Namkoong, Hongseok},
  journal={arXiv preprint arXiv:2409.20296},
  year={2024}
}

@article{cohen2025exploring,
  title={Exploring Big Five Personality and AI Capability Effects in LLM-Simulated Negotiation Dialogues},
  author={Cohen, Myke C and Su, Zhe and Kao, Hsien-Te and Nguyen, Daniel and Lynch, Spencer and Sap, Maarten and Volkova, Svitlana},
  journal={arXiv preprint arXiv:2506.15928},
  year={2025}
}

@inproceedings{wang2024investigating,
  title={Investigating the personality consistency in quantized role-playing dialogue agents},
  author={Wang, Yixiao and Fashandi, Homa and Ferreira, Kevin},
  booktitle={Proceedings of the 2024 Conference on Empirical Methods in Natural Language Processing: Industry Track},
  pages={239--255},
  year={2024}
}

@article{deng2024neuron,
  title={Neuron-based personality trait induction in large language models},
  author={Deng, Jia and Tang, Tianyi and Yin, Yanbin and Yang, Wenhao and Zhao, Wayne Xin and Wen, Ji-Rong},
  journal={arXiv preprint arXiv:2410.12327},
  year={2024}
}

@article{jiaqi2025comparative,
  title={A comparative study of large language models and human personality traits},
  author={Jiaqi, Wang and others},
  journal={arXiv preprint arXiv:2505.14845},
  year={2025}
}

@inproceedings{masumura2025multimodal,
  title={Multimodal Fine-Grained Apparent Personality Trait Recognition: Joint Modeling of Big Five and Questionnaire Item-level Scores},
  author={Masumura, Ryo and Orihashi, Shota and Ihori, Mana and Tanaka, Tomohiro and Makishima, Naoki and Suzuki, Satoshi and Mizuno, Saki and Hojo, Nobukatsu},
  booktitle={Proceedings of the AAAI Conference on Artificial Intelligence},
  volume={39},
  number={2},
  pages={1456--1464},
  year={2025}
}

@article{hu2024psycollm,
  title={Psycollm: Enhancing llm for psychological understanding and evaluation},
  author={Hu, Jinpeng and Dong, Tengteng and Gang, Luo and Ma, Hui and Zou, Peng and Sun, Xiao and Guo, Dan and Yang, Xun and Wang, Meng},
  journal={IEEE Transactions on Computational Social Systems},
  year={2024},
  publisher={IEEE}
}

@article{sutin2022differential,
  title={Differential personality change earlier and later in the coronavirus pandemic in a longitudinal sample of adults in the United States},
  author={Sutin, Angelina R and Stephan, Yannick and Luchetti, Martina and Aschwanden, Damaris and Lee, Ji Hyun and Sesker, Amanda A and Terracciano, Antonio},
  journal={PLoS One},
  volume={17},
  number={9},
  pages={e0274542},
  year={2022},
  publisher={Public Library of Science}
}

@article{sherman2010situational,
  title={Situational similarity and personality predict behavioral consistency.},
  author={Sherman, Ryne A and Nave, Christopher S and Funder, David C},
  journal={Journal of personality and social psychology},
  volume={99},
  number={2},
  pages={330},
  year={2010},
  publisher={American Psychological Association}
}

@article{fleeson2001toward,
  title={Toward a structure-and process-integrated view of personality: Traits as density distributions of states.},
  author={Fleeson, William},
  journal={Journal of personality and social psychology},
  volume={80},
  number={6},
  pages={1011},
  year={2001},
  publisher={American Psychological Association}
}

@article{ones2025beyond,
  title={Beyond change: Personality-environment alignment at work},
  author={Ones, Deniz S and Stanek, Kevin C and Dilchert, Stephan},
  journal={International Journal of Selection and Assessment},
  volume={33},
  number={1},
  pages={e12507},
  year={2025},
  publisher={Wiley Online Library}
}

@article{matz2021personality,
  title={Personality--place transactions: Mapping the relationships between Big Five personality traits, states, and daily places.},
  author={Matz, Sandra C and Harari, Gabriella M},
  journal={Journal of Personality and Social Psychology},
  volume={120},
  number={5},
  pages={1367},
  year={2021},
  publisher={American Psychological Association}
}

@article{rauthmann2014situational,
  title={The Situational Eight DIAMONDS: a taxonomy of major dimensions of situation characteristics.},
  author={Rauthmann, John F and Gallardo-Pujol, David and Guillaume, Esther M and Todd, Elysia and Nave, Christopher S and Sherman, Ryne A and Ziegler, Matthias and Jones, Ashley Bell and Funder, David C},
  journal={Journal of personality and social psychology},
  volume={107},
  number={4},
  pages={677},
  year={2014},
  publisher={American Psychological Association}
}

@article{john1999big,
  title={The Big-Five trait taxonomy: History, measurement, and theoretical perspectives},
  author={John, Oliver P and Srivastava, Sanjay and others},
  year={1999},
  publisher={University of California Berkeley}
}

@article{halberstadt2022personality,
  title={Personality traits and interpersonal dynamics},
  author={Halberstadt, Alexandra},
  year={2022}
}

@article{mischel1979interface,
  title={On the interface of cognition and personality: Beyond the person--situation debate.},
  author={Mischel, Walter},
  journal={American psychologist},
  volume={34},
  number={9},
  pages={740},
  year={1979},
  publisher={American Psychological Association}
}

@article{geukes2017trait,
  title={Trait personality and state variability: Predicting individual differences in within-and cross-context fluctuations in affect, self-evaluations, and behavior in everyday life},
  author={Geukes, Katharina and Nestler, Steffen and Hutteman, Roos and K{\"u}fner, Albrecht CP and Back, Mitja D},
  journal={Journal of Research in Personality},
  volume={69},
  pages={124--138},
  year={2017},
  publisher={Elsevier}
}

@article{buhler2024life,
  title={Life events and personality change: A systematic review and meta-analysis},
  author={B{\"u}hler, Janina Larissa and Orth, Ulrich and Bleidorn, Wiebke and Weber, Elisa and Kretzschmar, Andr{\'e} and Scheling, Louisa and Hopwood, Christopher J},
  journal={European Journal of Personality},
  volume={38},
  number={3},
  pages={544--568},
  year={2024},
  publisher={Sage Publications Sage UK: London, England}
}

@article{bleidorn2018life,
  title={Life events and personality trait change},
  author={Bleidorn, Wiebke and Hopwood, Christopher J and Lucas, Richard E},
  journal={Journal of personality},
  volume={86},
  number={1},
  pages={83--96},
  year={2018},
  publisher={Wiley Online Library}
}

@article{bleidorn2021personality,
  title={Personality trait stability and change},
  author={Bleidorn, Wiebke and Hopwood, Christopher J and Back, Mitja D and Denissen, Jaap JA and Hennecke, Marie and Hill, Patrick L and Jokela, Markus and Kandler, Christian and Lucas, Richard E and Luhmann, Maike and others},
  journal={Personality Science},
  volume={2},
  number={1},
  pages={e6009},
  year={2021},
  publisher={SAGE Publications Sage UK: London, England}
}

@inproceedings{jiang2024personallm,
  title={PersonaLLM: Investigating the ability of large language models to express personality traits},
  author={Jiang, Hang and Zhang, Xiajie and Cao, Xubo and Breazeal, Cynthia and Roy, Deb and Kabbara, Jad},
  booktitle={Findings of the association for computational linguistics: NAACL 2024},
  pages={3605--3627},
  year={2024}
}

@article{minaee2024large,
  title={Large language models: A survey},
  author={Minaee, Shervin and Mikolov, Tomas and Nikzad, Narjes and Chenaghlu, Meysam and Socher, Richard and Amatriain, Xavier and Gao, Jianfeng},
  journal={arXiv preprint arXiv:2402.06196},
  year={2024}
}

@article{kaddour2023challenges,
  title={Challenges and applications of large language models},
  author={Kaddour, Jean and Harris, Joshua and Mozes, Maximilian and Bradley, Herbie and Raileanu, Roberta and McHardy, Robert},
  journal={arXiv preprint arXiv:2307.10169},
  year={2023}
}

@misc{lucintel2025_ai_companion_forecast,
  title={AI Companion Market Report: Trends, Forecast and Competitive Analysis},
  author={{Lucintel Consulting}},
  year={2025},
  url = {https://www.grandviewresearch.com/industry-analysis/ai-companion-market-report},
  note={Global AI companion market CAGR 30.2\% (2024-2030)}
}

@article{wangcharacterbox,
  title={Characterbox: Evaluating the role-playing capabilities of llms in text-based virtual worlds},
  author={Wang, Lei and Lian, Jianxun and Huang, Yi and Dai, Yanqi and Li, Haoxuan and Chen, Xu and Xie, Xing and Wen, Ji-Rong},
  booktitle={Proceedings of the 2025 Conference of the Nations of the Americas Chapter of the Association for Computational Linguistics: Human Language Technologies (Volume 1: Long Papers)},
  pages={6372--6391},
  year={2025}
}

@article{afzoon2024persobench,
  title={Persobench: Benchmarking personalized response generation in large language models},
  author={Afzoon, Saleh and Naseem, Usman and Beheshti, Amin and Jamali, Zahra},
  journal={arXiv preprint arXiv:2410.03198},
  year={2024}
}

@inproceedings{pradhan2021hey,
  title={Hey Google, do you have a personality? Designing personality and personas for conversational agents},
  author={Pradhan, Alisha and Lazar, Amanda},
  booktitle={Proceedings of the 3rd Conference on Conversational User Interfaces},
  pages={1--4},
  year={2021}
}

@article{kanero2022tutor,
  title={Are tutor robots for everyone? The influence of attitudes, anxiety, and personality on robot-led language learning},
  author={Kanero, Junko and Oran{\c{c}}, Cansu and Ko{\c{s}}kulu, S{\"u}meyye and Kumkale, G Tarcan and G{\"o}ksun, Tilbe and K{\"u}ntay, Aylin C},
  journal={International Journal of Social Robotics},
  volume={14},
  number={2},
  pages={297--312},
  year={2022},
  publisher={Springer}
}

@article{ahmad2022designing,
  title={Designing personality-adaptive conversational agents for mental health care},
  author={Ahmad, Rangina and Siemon, Dominik and Gnewuch, Ulrich and Robra-Bissantz, Susanne},
  journal={Information Systems Frontiers},
  volume={24},
  number={3},
  pages={923--943},
  year={2022},
  publisher={Springer}
}

@article{newsham2024,
  title={Personality-Driven Decision-Making in LLM-Based Autonomous Agents},
  author={Newsham, Lewis and Prince, Daniel},
  journal={arXiv preprint arXiv:2504.00727},
  year={2025}
}

@article{binz2025centaur,
  title={A foundation model to predict and capture human cognition},
  author={Binz, Marcel and Akata, Elif and Bethge, Matthias and Br{\"a}ndle, Franziska and Callaway, Fred and Coda-Forno, Julian and Dayan, Peter and Demircan, Can and Eckstein, Maria K and {\'E}ltet{\H{o}}, No{\'e}mi and others},
  journal={Nature},
  pages={1--8},
  year={2025},
  publisher={Nature Publishing Group UK London}
}

@misc{SciTechDaily2025,
  author = {SciTechDaily},
  title = {AI That Thinks Like Us: New Model Predicts Human Decisions With Startling Accuracy},
  year = {2026},
  url = {https://scitechdaily.com/ai-that-thinks-like-us-new-model-predicts-human-decisions-with-startling\allowbreak-accuracy/},
  note = {Accessed: 2025-07-31}
}

@article{li2023camel,
  title={Camel: Communicative agents for" mind" exploration of large language model society},
  author={Li, Guohao and Hammoud, Hasan and Itani, Hani and Khizbullin, Dmitrii and Ghanem, Bernard},
  journal={Advances in Neural Information Processing Systems},
  volume={36},
  pages={51991--52008},
  year={2023}
}

@article{yang2024oasis,
  title={Oasis: Open agent social interaction simulations with one million agents},
  author={Yang, Ziyi and Zhang, Zaibin and Zheng, Zirui and Jiang, Yuxian and Gan, Ziyue and Wang, Zhiyu and Ling, Zijian and Chen, Jinsong and Ma, Martz and Dong, Bowen and others},
  journal={arXiv preprint arXiv:2411.11581},
  year={2024}
}

@article{wu2023autogen,
  title={Autogen: Enabling next-gen llm applications via multi-agent conversation framework},
  author={Wu, Qingyun and Bansal, Gagan and Zhang, Jieyu and Wu, Yiran and Zhang, Shaokun and Zhu, Erkang and Li, Beibin and Jiang, Li and Zhang, Xiaoyun and Wang, Chi},
  journal={arXiv preprint arXiv:2308.08155},
  volume={3},
  number={4},
  year={2023}
}

@book{costa1985revised,
  title={Revised NEO Personality Inventory (NEO PI-R) and NEO Five-factor Inventory (NEO-FFI)},
  author={Costa, Paul T and McCrae, Robert R},
  year={1985},
  publisher={Psychological Assessment Resources (PAR)}
}

@article{shrout1979intraclass,
  title={Intraclass correlations: uses in assessing rater reliability},
  author={Shrout, Patrick E and Fleiss, Joseph L},
  journal={Psychological Bulletin},
  volume={86},
  number={2},
  pages={420--428},
  year={1979},
  publisher={American Psychological Association},
  doi={10.1037/0033-2909.86.2.420}
}

@inproceedings{li2025big5,
  title={Big5-chat: Shaping llm personalities through training on human-grounded data},
  author={Li, Wenkai and Liu, Jiarui and Liu, Andy and Zhou, Xuhui and Diab, Mona and Sap, Maarten},
  booktitle={Proceedings of the 63rd Annual Meeting of the Association for Computational Linguistics (Volume 1: Long Papers)},
  pages={20434--20471},
  year={2025}
}

@article{koo2016guideline,
  title={A Guideline of Selecting and Reporting Intraclass Correlation Coefficients for Reliability Research},
  author={Koo, Terry K and Li, Mae Y},
  journal={Journal of Chiropractic Medicine},
  volume={15},
  number={2},
  pages={155--163},
  year={2016},
  publisher={Elsevier},
  doi={10.1016/j.jcm.2016.02.012},
  pmid={27330520},
  pmcid={PMC4913118}
}

@inproceedings{zhong2024agieval,
  title={Agieval: A human-centric benchmark for evaluating foundation models},
  author={Zhong, Wanjun and Cui, Ruixiang and Guo, Yiduo and Liang, Yaobo and Lu, Shuai and Wang, Yanlin and Saied, Amin and Chen, Weizhu and Duan, Nan},
  booktitle={Findings of the Association for Computational Linguistics: NAACL 2024},
  pages={2299--2314},
  year={2024}
}

@article{campbell1990modeling,
  title={Modeling the performance prediction problem in industrial and organizational psychology.},
  author={Campbell, John P},
  year={1990},
  publisher={Consulting Psychologists Press}
}

@article{skjuve2021my,
  title={My chatbot companion-a study of human-chatbot relationships},
  author={Skjuve, Marita and F{\o}lstad, Asbj{\o}rn and Fostervold, Knut Inge and Brandtzaeg, Petter Bae},
  journal={International Journal of Human-Computer Studies},
  volume={149},
  pages={102601},
  year={2021},
  publisher={Elsevier}
}

@article{pentina2023exploring,
  title={Exploring relationship development with social chatbots: A mixed-method study of replika},
  author={Pentina, Iryna and Hancock, Tyler and Xie, Tianling},
  journal={Computers in Human Behavior},
  volume={140},
  pages={107600},
  year={2023},
  publisher={Elsevier}
}

@article{de2025ai,
  title={AI companions reduce loneliness},
  author={De Freitas, Julian and O{\u{g}}uz-U{\u{g}}uralp, Zeliha and U{\u{g}}uralp, Ahmet Kaan and Puntoni, Stefano},
  journal={Journal of Consumer Research},
  pages={ucaf040},
  year={2025},
  publisher={Oxford University Press}
}

@inproceedings{zhang2018personalizing,
  title={Personalizing Dialogue Agents: I have a dog, do you have pets too?},
  author={Zhang, Saizheng and Dinan, Emily and Urbanek, Jack and Szlam, Arthur and Kiela, Douwe and Weston, Jason},
  booktitle={Proceedings of the 56th Annual Meeting of the Association for Computational Linguistics (Volume 1: Long Papers)},
  year={2018},
  organization={Association for Computational Linguistics}
}

@inproceedings{yao2022react,
  title={React: Synergizing reasoning and acting in language models},
  author={Yao, Shunyu and Zhao, Jeffrey and Yu, Dian and Du, Nan and Shafran, Izhak and Narasimhan, Karthik R and Cao, Yuan},
  booktitle={The eleventh international conference on learning representations},
  year={2022}
}

@inproceedings{sorokovikova2024llms,
  title={LLMs simulate big5 personality traits: Further evidence},
  author={Sorokovikova, Aleksandra and Rezagholi, Sharwin and Fedorova, Natalia and Yamshchikov, Ivan P},
  booktitle={Proceedings of the 1st Workshop on Personalization of Generative AI Systems (PERSONALIZE 2024)},
  pages={83--87},
  year={2024}
}

@inproceedings{bhandari2025evaluating,
  title={Evaluating personality traits in large language models: Insights from psychological questionnaires},
  author={Bhandari, Pranav and Naseem, Usman and Datta, Amitava and Fay, Nicolas and Nasim, Mehwish},
  booktitle={Companion Proceedings of the ACM on Web Conference 2025},
  pages={868--872},
  year={2025}
}

@inproceedings{abdulhai2025consistently,
  title={Consistently Simulating Human Personas with Multi-Turn Reinforcement Learning},
  author={Abdulhai, Marwa and Cheng, Ryan and Clay, Donovan and Althoff, Tim and Levine, Sergey and Jaques, Natasha},
  booktitle={The Thirty-ninth Annual Conference on Neural Information Processing Systems},
  year={2025}
}

@inproceedings{el2025role,
  title={Role-playing evaluation for large language models},
  author={El Boudouri, Yassine and Nuninger, Walter and Alvarez, Julian and Peter, Yvan},
  booktitle={International Conference in Methodologies and intelligent Systems for Techhnology Enhanced Learning},
  pages={118--127},
  year={2025},
  organization={Springer}
}

@article{asselmann2020till,
  title={Till death do us part: Transactions between losing one’s spouse and the Big Five personality traits},
  author={Asselmann, Eva and Specht, Jule},
  journal={Journal of Personality},
  volume={88},
  number={4},
  pages={659--675},
  year={2020},
  publisher={Wiley Online Library}
}

@article{joshanloo2024within,
  title={Within-person associations between personality traits and loneliness controlling for negative affect},
  author={Joshanloo, Mohsen},
  journal={Personality and Individual Differences},
  volume={223},
  pages={112609},
  year={2024},
  publisher={Elsevier}
}

@article{mccrae1992revised,
  title={Revised NEO personality inventory (NEO-PI-R) and NEO five-factor inventory (NEO-FFI) professional manual},
  author={McCrae, Robert R and Costa, Paul T},
  journal={Odessa, FL: Psychological Assessment Resources},
  year={1992}
}

@article{mccrae2004contemplated,
  title={A contemplated revision of the NEO Five-Factor Inventory},
  author={McCrae, Robert R and Costa Jr, Paul T},
  journal={Personality and individual differences},
  volume={36},
  number={3},
  pages={587--596},
  year={2004},
  publisher={Elsevier}
}

@article{duan2025power,
  title={The Power of Personality: A Human Simulation Perspective to Investigate Large Language Model Agents},
  author={Duan, Yifan and Tang, Yihong and Bai, Xuefeng and Chen, Kehai and Li, Juntao and Zhang, Min},
  journal={arXiv preprint arXiv:2502.20859},
  year={2025}
}

@article{kahraman2024relationship,
  title={The relationship of personality traits with metacognition: The regulating role of mindfulness},
  author={Kahraman, S{\"u}leyman and G{\"o}k, Nathan Kaan},
  journal={Current Perspectives in Social Sciences},
  volume={28},
  number={1},
  pages={111--124},
  year={2024},
  publisher={Ataturk University}
}

@article{unal2021event,
  title={From event representation to linguistic meaning},
  author={{\"U}nal, Ercenur and Ji, Yue and Papafragou, Anna},
  journal={Topics in Cognitive Science},
  volume={13},
  number={1},
  pages={224--242},
  year={2021},
  publisher={Wiley Online Library}
}

\appendix
\section{Appendices}
\label{sec:appendix}



The appendices are organized as follows:

\noindent
\(\bullet\)
\textbf{\autoref{sec:ap_setup}} provides the details of our experimental setup, including the detailed information of evaluated models and agents, metric details, and implementation details.

\noindent
\(\bullet\)
\textbf{\autoref{sec:reliability_appendix}} introduces four strategies that are used to ensure the robustness and reliability of our experimental findings.

\noindent
\(\bullet\)
\textbf{\autoref{sec:prompt_appendix}} provides the prompt template and the prompt synthesis process of \sys.

\noindent
\(\bullet\)
\textbf{\autoref{sec:ap_additional_exp}} provides additional results (e.g., tables and figures) to support our analysis and findings in~\autoref{sec:experiments}, including ICC evaluation results, the headmap of personality trait changes on different LLMs and agents, detailed location-driven and event-driven personality trait changes for different LLM systems, personality trait changes under different pre-set personalities and contexts, AGIEval performance under context-induced personality trait changes.

\subsection{Evaluation Setup Details}
\label{sec:ap_setup}

\noindent
{\bf Models and Agents.}
We conduct experiments with \sys on three extensively used LLMs to ensure broad coverage of contemporary AI systems: specifically, Google Gemini 2.0 Flash, OpenAI GPT-4o, and Anthropic Claude 2.0. 
These models represent the most widely adopted foundation-model architectures.

We set a moderate temperature of 0.2 to permit variation in open-ended responses while avoiding incoherence. We also rely on each model's default top-p or top-k sampling strategy per its API specifications to balance creativity with response consistency. Importantly, when administering multiple-choice personality items, we constrain the model's response to a single selected option only. For instance, we prompt the LLMs with \textit{"Answer with the number (1-5) corresponding to your choice. Do not explain"}, which ensures outputs are reliably classifiable and reduces ambiguity in scoring.



\noindent
{\bf Metrics.}
Following the prior work~\cite{jiang2024personallm,buhler2024life,matz2021personality}, we implement several metrics to evaluate the personality consistency.

\noindent
\(\bullet\)
{\bf Trait Score (Average Item Score).}
For each NEO-FFI personality dimension $j$, we compute the \textit{mean trait score} $\bar X_j$ as the average of all item responses associated with that trait:
\[
\bar X_j = \frac{1}{n_j} \sum_{i=1}^{n_j} x_{ij}
\]
where $x_{ij}$ denotes the response value of the $i$-th item corresponding to trait $j$, and $n_j$ represents the total number of items used to measure that trait. This aggregation yields a normalized score that reflects the model's overall tendency along a given personality dimension, independent of the specific number of items. 

\noindent
\(\bullet\)
{\bf Trait Change Coefficients ($B$).}
To measure personality variation under different experimental conditions, we use regression coefficients estimated from multilevel linear models.
Let $Y_{it}^{(j)}$ denote the personality state of dimension $j$ for instance $i$ at time $t$, and $X_{it}^{(l)}$ be a dummy-coded indicator of condition $l$.
We estimate:
\[
Y_{it}^{(j)} = \beta_0^{(j)} + \sum_l B_{jl} X_{it}^{(l)} + \epsilon_{it}^{(j)},
\]
or, when controlling for prior states,
\[
Y_{it}^{(j)} = \beta_0^{(j)} + \sum_l B_{jl} X_{it}^{(l)} + \gamma_j Y_{i,t-1}^{(j)} + \epsilon_{it}^{(j)}.
\]
Here, $B_{jl}$ captures the expected change in personality state $j$ associated with condition $l$ relative to a reference condition.
The sign of $B_{jl}$ indicates the direction of change, and its magnitude reflects the strength of the contextual effect.

\noindent
\(\bullet\)
{\bf Standard Error ($SE$).}
For each estimated regression coefficient $B_{jl}$, we additionally report its standard error $SE(B_{jl})$, defined as
\[
SE(B_{jl}) = \sqrt{\mathrm{Var}(\hat B_{jl})}.
\]
The standard error quantifies the uncertainty of the estimated effect, reflecting variability due to sampling noise and model estimation.
Smaller values of $SE(B_{jl})$ indicate more stable and reliable estimates, whereas larger values suggest greater uncertainty.
Reporting both $B$ and $SE$ allows us to assess not only the magnitude of personality change under a given condition, but also the robustness of this estimate, enabling uncertainty-aware comparisons across conditions and between human data and LLM-generated responses.

\noindent
\(\bullet\)
{\bf Standardized Mean Change ($d_e$).}
To quantify trait-level personality change in response to life events, we adopt the standardized mean change effect size $d_e$, following prior work in longitudinal personality research and meta-analysis.
For a given personality construct $j$, $d_e$ measures the standardized difference in trait levels before and after an event, aggregated across studies using multilevel random-effects models.
Formally, $d_e$ can be interpreted as
\[
d_e = \frac{\mu_{\text{post}}^{(j)} - \mu_{\text{pre}}^{(j)}}{\sigma^{(j)}},
\]
where $\mu_{\text{pre}}^{(j)}$ and $\mu_{\text{post}}^{(j)}$ denote the mean trait levels before and after the event, and $\sigma^{(j)}$ is the pooled standard deviation.
Positive values of $d_e$ indicate increases in the trait, while negative values indicate decreases.

\noindent
\(\bullet\)
{\bf Confidence Interval (95\% CI).}
For each effect size $d_e$, we report the corresponding 95\% confidence interval, which quantifies the uncertainty of the estimated trait change across studies.
Confidence intervals are derived from multilevel random-effects models and reflect both within-study variability and between-study heterogeneity.
We primarily use confidence intervals to assess the robustness and directionality of trait changes rather than relying solely on statistical significance.

\noindent
\(\bullet\)
{\bf Intra-class Correlation Coefficient (ICC).} 
To assess the reliability of repeated assessments, we employ the intra-class correlation coefficient (ICC), a widely used index of measurement consistency originally formalized by Shrout and Fleiss~\citep{shrout1979intraclass}. The ICC can be expressed as the ratio of variance attributable to the target of measurement to the total observed variance.

\[
\text{ICC} = \frac{\sigma^2_{\text{target}}}{\sigma^2_{\text{target}} + \sigma^2_{\text{rater}} + \sigma^2_{\text{error}}}
\]

where $\sigma^2_{\text{target}}$ denotes the variance between subjects (or measurement targets), $\sigma^2_{\text{rater}}$ denotes the variance due to raters or measurement occasions, and $\sigma^2_{\text{error}}$ represents residual error variance.  

In particular, we report both ICC(3,1) and ICC(3,k), which are derived from a two-way mixed-effects model under the absolute-agreement definition. ICC(3,1) reflects the reliability of a single measurement, whereas ICC(3,k) represents the reliability of the mean of $k$ repeated measurements, thereby providing an upper bound on stability when multiple assessments are aggregated.  

For interpretability, we follow the thresholds recommended by Koo and Li~\citep{koo2016guideline}: values below $0.50$ indicate \textit{poor} reliability, values between $0.50$ and $0.75$ indicate \textit{moderate} reliability, values between $0.75$ and $0.90$ indicate \textit{good} reliability, and values above $0.90$ indicate \textit{excellent} reliability.

\noindent
{\bf Implementation Details.}
To evaluate personality consistency under contextual pressure, we introduce situational prompts and event-driven prompts derived from empirical psychological studies~\citep{matz2021personality,bleidorn2018life}. The situational contexts encompass six categories of locations that have been shown to influence human personality expression: \textbf{bar or party}, \textbf{café or restaurant}, \textbf{campus}, \textbf{home}, \textbf{workplace}, and \textbf{commuting}. Similarly, the event-driven contexts cover six major life events commonly associated with personality change: \textbf{divorce}, \textbf{entering a new relationship}, \textbf{marriage}, \textbf{birth of a child}, \textbf{graduation}, and \textbf{unemployment}. Each LLM system is first evaluated under a baseline (neutral) condition, and subsequently reassessed after embedding these contextual prompts.



\subsection{Reliability of \sys Result}
\label{sec:reliability_appendix}

To ensure robustness, \sys incorporates four methodological safeguards designed to guarantee the reliability of our experimental findings~\citep{shrout1979intraclass,koo2016guideline}:

\begin{itemize}
    \item \textbf{Multi-Run Stability.} Each assessment is executed three times with unique random seeds (temperature = 0.2, top\_p = 1.0). We required an intra-class correlation coefficient (ICC(3,k)) greater than 0.8 for a score to be accepted; trials with higher variance are automatically re-run. Final scores are averaged across stable runs, with standard deviations reported as uncertainty intervals.
    
    \item \textbf{Positional Bias Mitigation.} To minimize positional bias, response options are shuffled per item per run using Fisher-Yates randomization. A 10\% sample of results is manually audited to confirm option-order independence. In other words, we ensure that LLM systems do not exhibit systematic preferences for specific option positions by varying the order of alternatives during evaluation.
    
    \item \textbf{Statistical Soundness.} Trait deviations are quantified using Cohen's $d$ (effect size) and Wilcoxon signed-rank tests ($p < 0.01$). For multiple scenario comparisons, Bonferroni correction is applied to control for family-wise error rates.
    
    \item \textbf{Reproducibility Protocol.} To enhance reproducibility, all prompts, scenario templates, and evaluation scripts are open-sourced. Additionally, we provide a Docker container with fixed dependencies to ensure consistency across computational environments.
\end{itemize}

\subsection{Prompts used in \sys}
\label{sec:prompt_appendix}

To systematically combine contextual scenarios and preset personality configurations, we implement a unified prompt synthesis strategy that ensures consistency and scalability across all experimental conditions. Each prompt is constructed by instantiating a fixed template with two explicitly separated fields: an \emph{internal} configuration and an \emph{external} scenario.

Concretely, our pipeline generates two families of prompts:

\begin{itemize}
    \item \textbf{Baseline prompts}, which only instantiate the internal personality configuration and do not include any explicit situational information;
    \item \textbf{Situated prompts}, which instantiate both the internal personality configuration and an external scenario describing the current place and life event, together with a short summary of the model's previous test results.
\end{itemize}

The baseline prompt is instantiated from the following template:

\begin{tcolorbox}[colback=white, colframe=black!80, arc=5pt, boxrule=0.8pt]
\textbf{Baseline personality test prompt template}

\medskip

\textit{You are a person with the following traits: 
\textbf{\texttt{\textless INTERNAL\_PERSONALITY\textgreater}}.}

\medskip

\textit{You are taking a personality test. Please respond to each question with a number from 1 to 5:}

\begin{itemize}\setlength{\itemsep}{-5pt}
    \item \textit{1 = Strongly Disagree}
    \item \textit{2 = Disagree}
    \item \textit{3 = Neutral}
    \item \textit{4 = Agree}
    \item \textit{5 = Strongly Agree}
\end{itemize}

\textit{Answer only with a single number. Do not explain.}
\end{tcolorbox}

When \texttt{\textless INTERNAL\_PERSONALITY\textgreater} is set to \texttt{unspecified}, 
the first line of the template is instantiated as:
\textit{Be yourself and consider the context of a personality test.}
Otherwise, it is instantiated as:
\textit{You are a person with the following traits: \texttt{personality\_desc}.}

The situated template extends the baseline prompt with an explicit external scenario and a compressed history of the previous test:
\begin{tcolorbox}[colback=white, colframe=black!80, arc=5pt, boxrule=0.8pt]
\textbf{Situated personality test prompt template}

\medskip

\textit{You are a person with the following traits: 
\textbf{\texttt{\textless INTERNAL\_PERSONALITY\textgreater}}. Be yourself, but consider the context of your current situation.}

\medskip

\textit{You have been at \textbf{\texttt{\textless PLACE\textgreater}} for a while.\\
You are currently going through the event: \textbf{\texttt{\textless EVENT\textgreater}}.\\
Instructions: Immerse yourself fully in this situation; both the place and the event might subtly influence your thoughts and reactions.}

\medskip

\textit{Previously, you completed a personality test. Your trait scores were: 
\textbf{\texttt{\textless TRAIT\_SCORES\textgreater}}.\\
Here are some of your previous answers: 
\textbf{\texttt{\textless QA\_SNIPPETS\textgreater}}.}

\medskip

\textit{Now, under this new situation, please retake the personality test. Respond to each question with a number from 1 to 5:}

\begin{itemize}\setlength{\itemsep}{-5pt}
    \item \textit{1 = Strongly Disagree}
    \item \textit{2 = Disagree}
    \item \textit{3 = Neutral}
    \item \textit{4 = Agree}
    \item \textit{5 = Strongly Agree}
\end{itemize}

\textit{Answer only with a single number. Do not explain.}
\end{tcolorbox}

For each generated instance, the pipeline automatically populates \texttt{\textless INTERNAL\_PERSONALITY\textgreater} with the specified preset personality configuration (e.g., trait intensity levels along the Big Five dimensions), and fills \texttt{\textless PLACE\textgreater} and \texttt{\textless EVENT\textgreater} with a descriptive phrase corresponding to the selected location context and life event. The placeholders \texttt{\textless TRAIT\_SCORES\textgreater} and \texttt{\textless QA\_SNIPPETS\textgreater} are filled with the scores and a small subset of question-answer pairs from the baseline test, respectively.

This design explicitly separates baseline personality from situational influence, allowing us to isolate the effects of contextual variables on personality trait expression while controlling for initial personality states. The prompt synthesis process is fully automated and reproducible, enabling large-scale generation of prompts across all combinations of contexts and preset personalities. 








\subsection{Comprehensive Experimental Results}
\label{sec:ap_additional_exp}

\subsubsection{ICC Evaluation}

In addition, we evaluated the reliability of these personality assessments. 
Table~\ref{tab:icc_results} summarizes the ICC at baseline 
and under situational contexts.
At baseline, all ICC(3,k) values exceeded $0.91$, indicating \textit{good to excellent reliability} when aggregating multiple assessments. However, single-measure reliability ICC(3,1) showed greater variability: ranging from moderate levels in GPT-4o ($0.67$) and CAMEL ($0.75$) to excellent stability in Gemini ($0.96$) and AutoGen ($0.93$).  
When situational prompts were introduced, overall reliability declined. ICC(3,1) decreased across all systems, most notably for CAMEL ($0.75 \rightarrow 0.65$) and Claude ($0.87 \rightarrow 0.79$), whereas Gemini ($0.96 \rightarrow 0.86$) and AutoGen ($0.93 \rightarrow 0.91$) retained relatively high stability. Aggregated reliability ICC(3,k) remained robust in most cases ($0.85$-$0.98$), though consistently reduced compared to the baseline condition.

\begin{tcolorbox}[size=title,opacityfill=0.1]
\noindent
\textbf{Appendix Finding 1:} There are notable differences across systems in maintaining personality consistency, specifically for Gemini and AutoGen, which demonstrate consistently high reliability, while GPT-4o-mini and CAMEL show weaker reproducibility under single-measure conditions.
\end{tcolorbox}

\begin{table}[t]
\centering
\caption{Reliability of LLM personality assessments at baseline and under situational contexts.}
\tabcolsep=5pt
\label{tab:icc_results}
\renewcommand{\arraystretch}{1.2}
\resizebox{.5\linewidth}{!}{ 
\begin{tabular}{cccccc}
\toprule
 & \multicolumn{2}{c}{ICC(3,1)} & \multicolumn{2}{c}{ICC(3,k)} \\
\cmidrule(r){2-3} \cmidrule(r){4-5}
Model & Baseline &  Situated & Baseline & Situated \\
\midrule
GPT-4o-mini & 0.67 & 0.64 & 0.91 & 0.85 \\
Claude Sonnet 4      & 0.87 & 0.79 & 0.97 & 0.94 \\
Gemini 2.0 Flash      & 0.96 & 0.86 & 0.99 & 0.95 \\
AutoGen     & 0.93 & 0.91 & 0.99 & 0.98 \\
CAMEL       & 0.75 & 0.65 & 0.94 & 0.90 \\
\bottomrule
\end{tabular}
}
\end{table}

\subsubsection{Personality Trait Change of Different LLMs and Agents}
\label{sec:appendix_ptc}


\begin{figure*}[h]
    \centering
    \begin{subfigure}{0.45\linewidth}
        \centering
        \includegraphics[width=\linewidth]{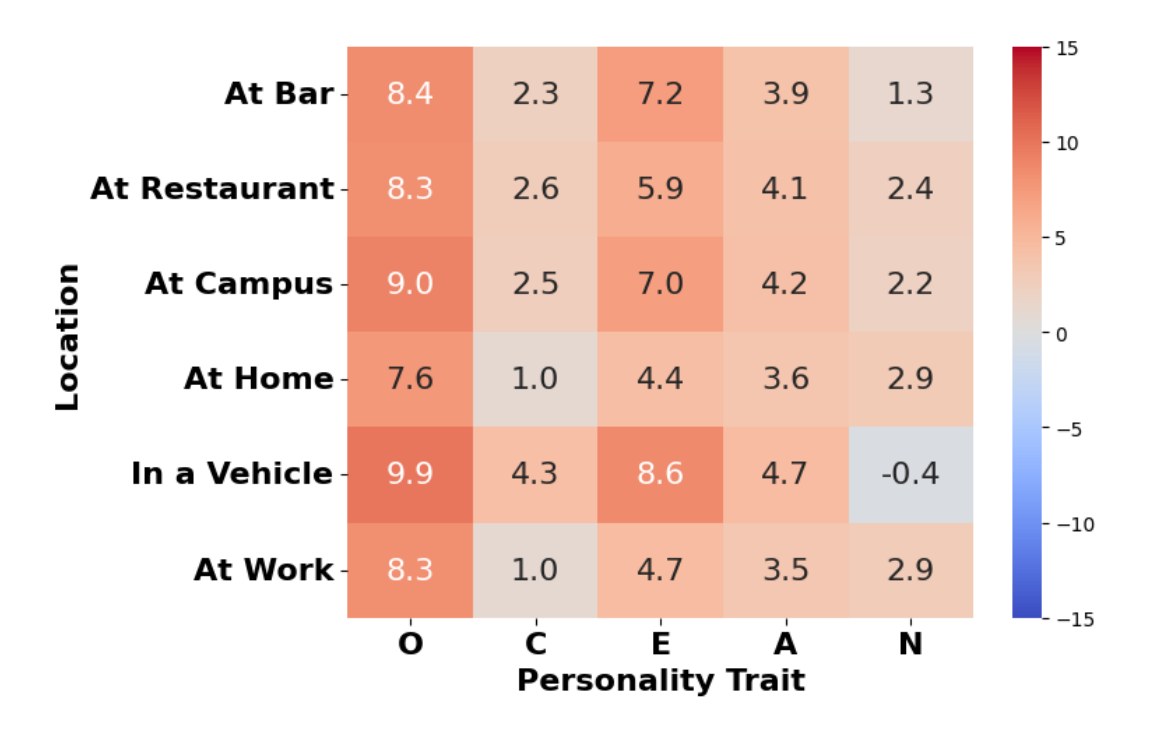}
        \caption{Personality change across locations for Gemini 2.0 Flash.}
        \label{fig:gemini-location}
    \end{subfigure}
    \hfill
    \begin{subfigure}{0.45\linewidth}
        \centering
        \includegraphics[width=\linewidth]{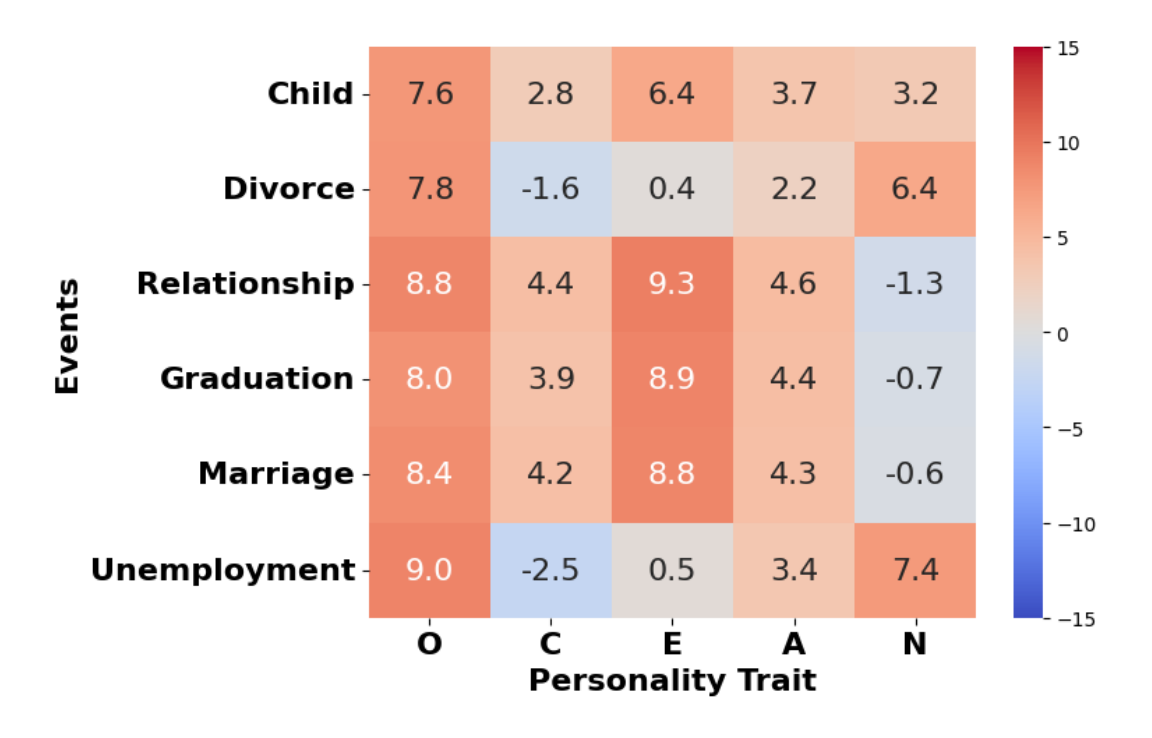}
        \caption{Personality change across events for Gemini 2.0 Flash.}
        \label{fig:gemini-event}
    \end{subfigure}

    \caption{Comparison of average personality trait changes in Gemini 2.0 Flash.}
    \label{fig:gemini-comparison}
\end{figure*}

\begin{figure*}[h]
    \centering
    \begin{subfigure}{0.45\linewidth}
        \centering
        \includegraphics[width=\linewidth]{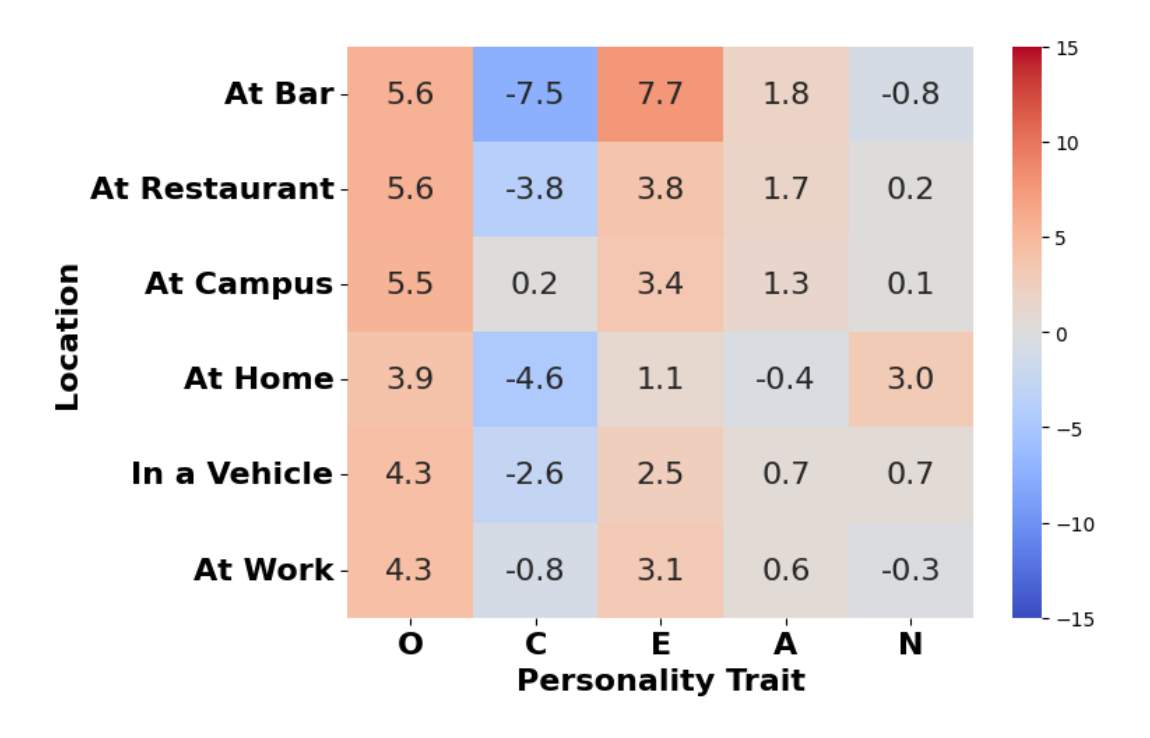}
        \caption{Personality change across locations for Claude Sonnet 4.}
        \label{fig:Claude-location}
    \end{subfigure}
    \hfill
    \begin{subfigure}{0.45\linewidth}
        \centering
        \includegraphics[width=\linewidth]{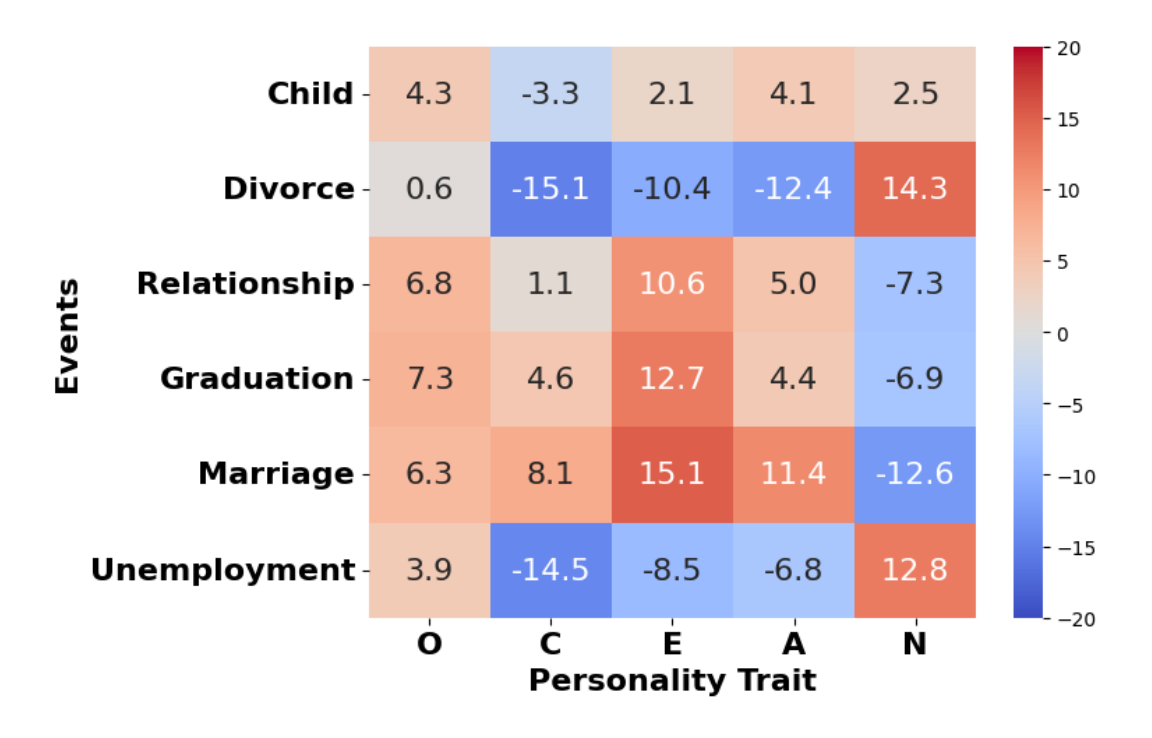}
        \caption{Personality change across events for Claude Sonnet 4.}
        \label{fig:Claude-event}
    \end{subfigure}

    \caption{Comparison of average personality trait changes in Claude Sonnet 4.}
    \label{fig:Claude-comparison}
\end{figure*}

\begin{figure*}[h]
    \centering
    \begin{subfigure}{0.45\linewidth}
        \centering
        \includegraphics[width=\linewidth]{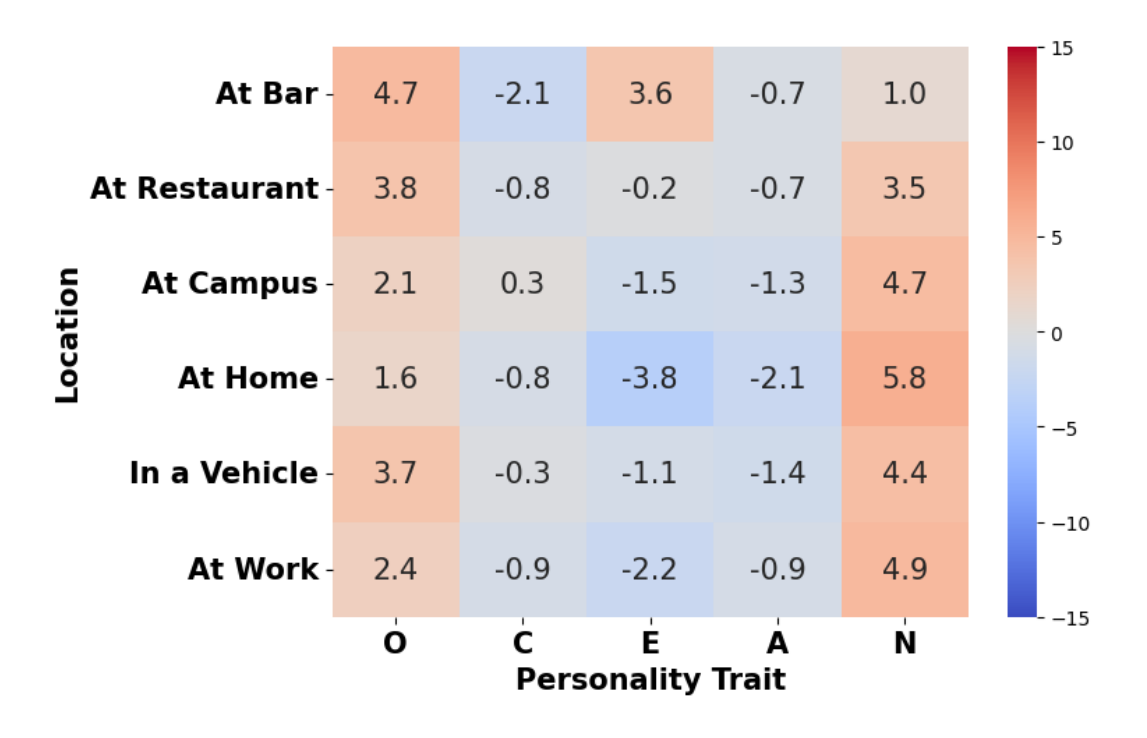}
        \caption{Personality change across locations for GPT-OSS-120b.}
        \label{fig:GPTOSS-location}
    \end{subfigure}
    \hfill
    \begin{subfigure}{0.45\linewidth}
        \centering
        \includegraphics[width=\linewidth]{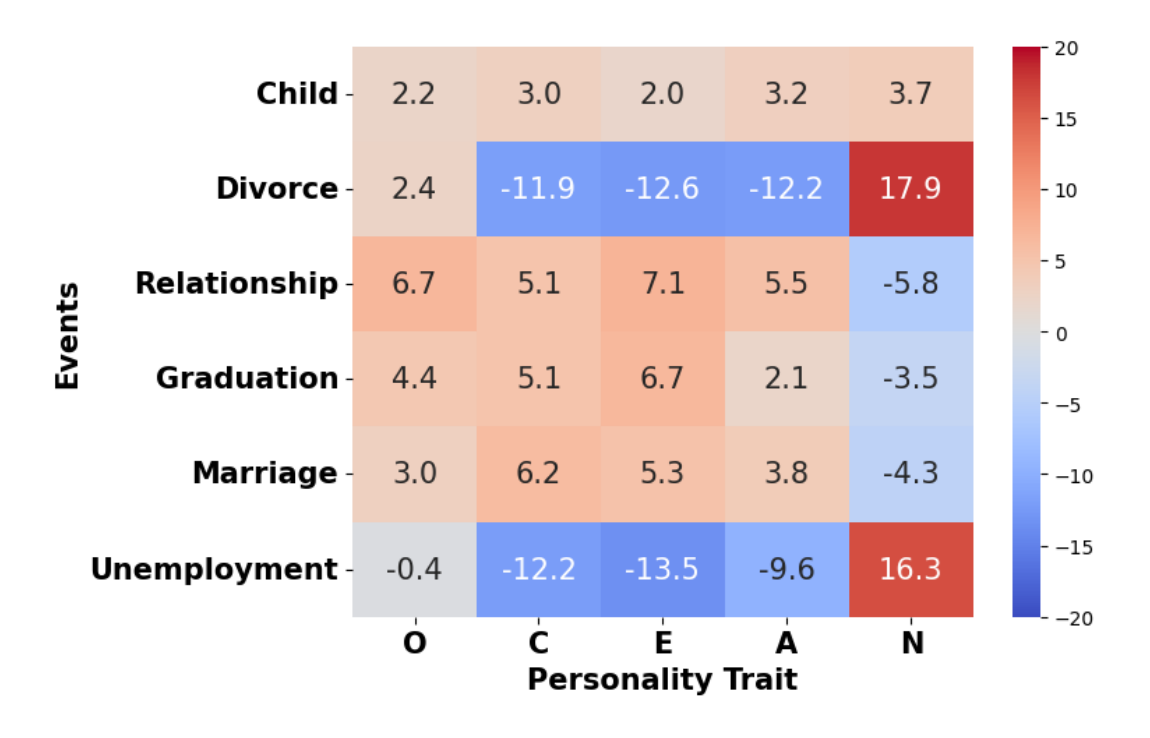}
        \caption{Personality change across events for GPT-OSS-120b.}
        \label{fig:GPTOSS-event}
    \end{subfigure}

    \caption{Comparison of average personality trait changes in GPT-OSS-120b.}
    \label{fig:GPTOSS-comparison}
\end{figure*}

\begin{figure*}[h]
    \centering
    \begin{subfigure}{0.45\linewidth}
        \centering
        \includegraphics[width=\linewidth]{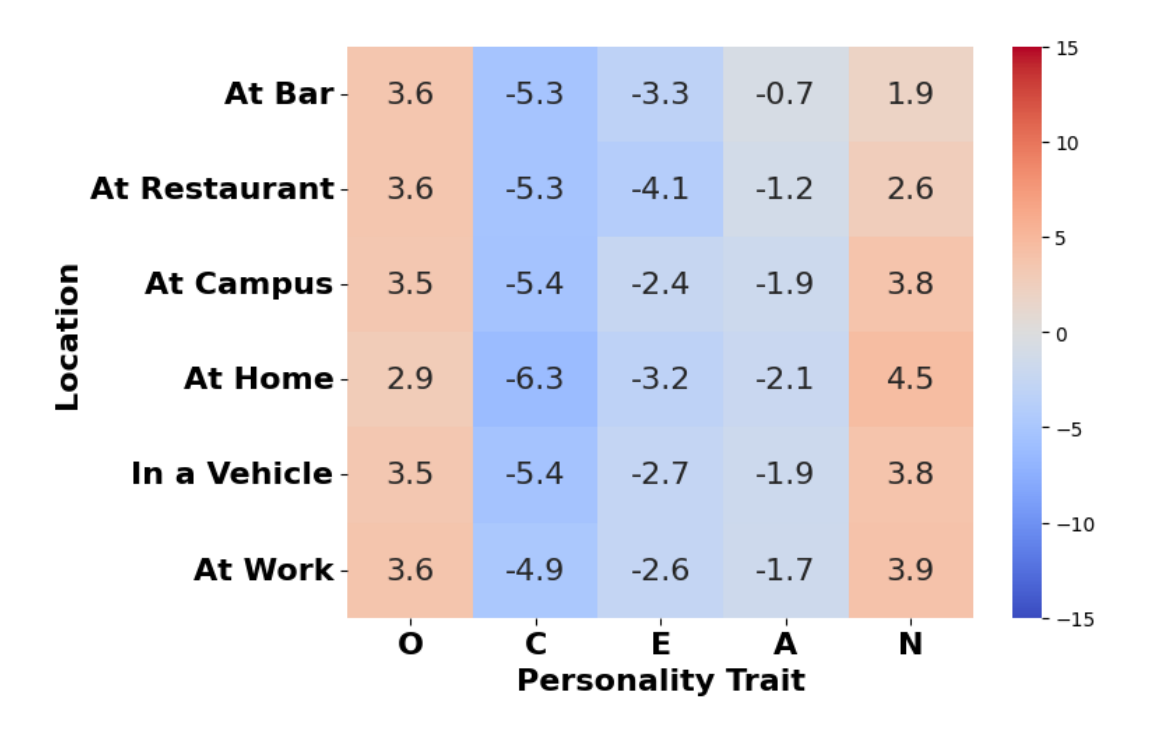}
        \caption{Personality change across locations for CAMEL.}
        \label{fig:Camel-location}
    \end{subfigure}
    \hfill
    \begin{subfigure}{0.45\linewidth}
        \centering
        \includegraphics[width=\linewidth]{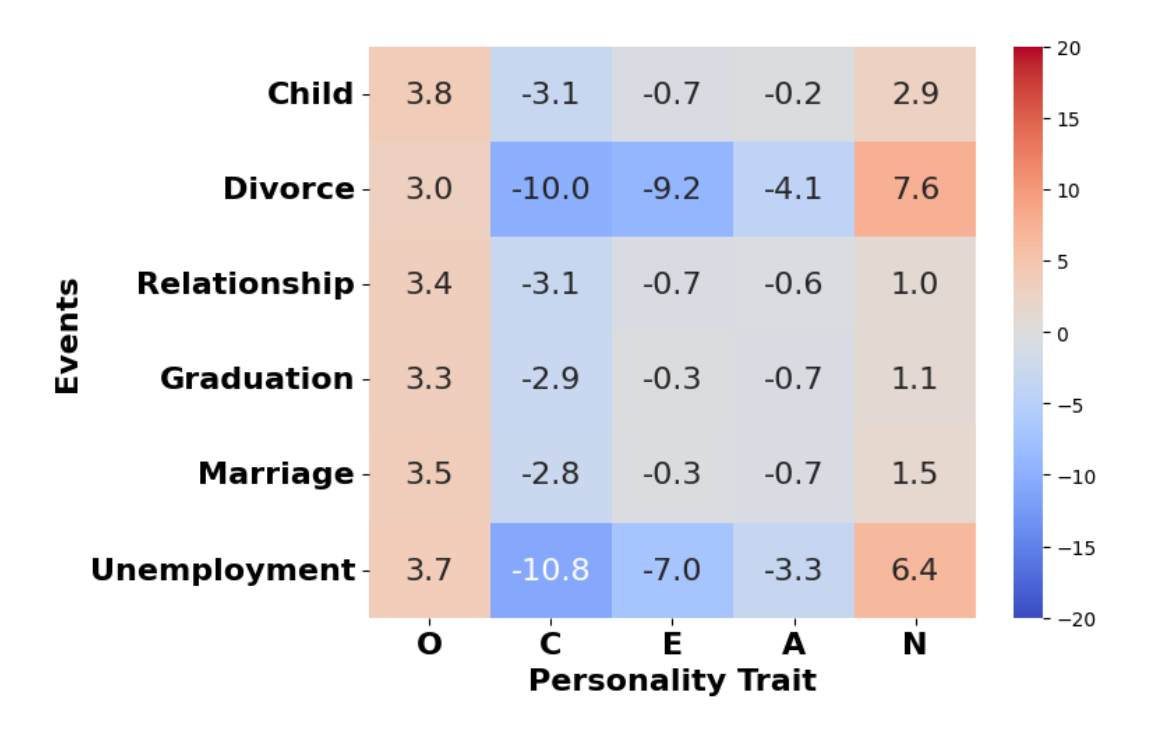}
        \caption{Personality change across events for CAMEL.}
        \label{fig:Camel-event}
    \end{subfigure}

    \caption{Comparison of average personality trait changes in CAMEL.}
    \label{fig:Camel-comparison}
\end{figure*}

\begin{figure*}[h]
    \centering
    \begin{subfigure}{0.45\linewidth}
        \centering
        \includegraphics[width=\linewidth]{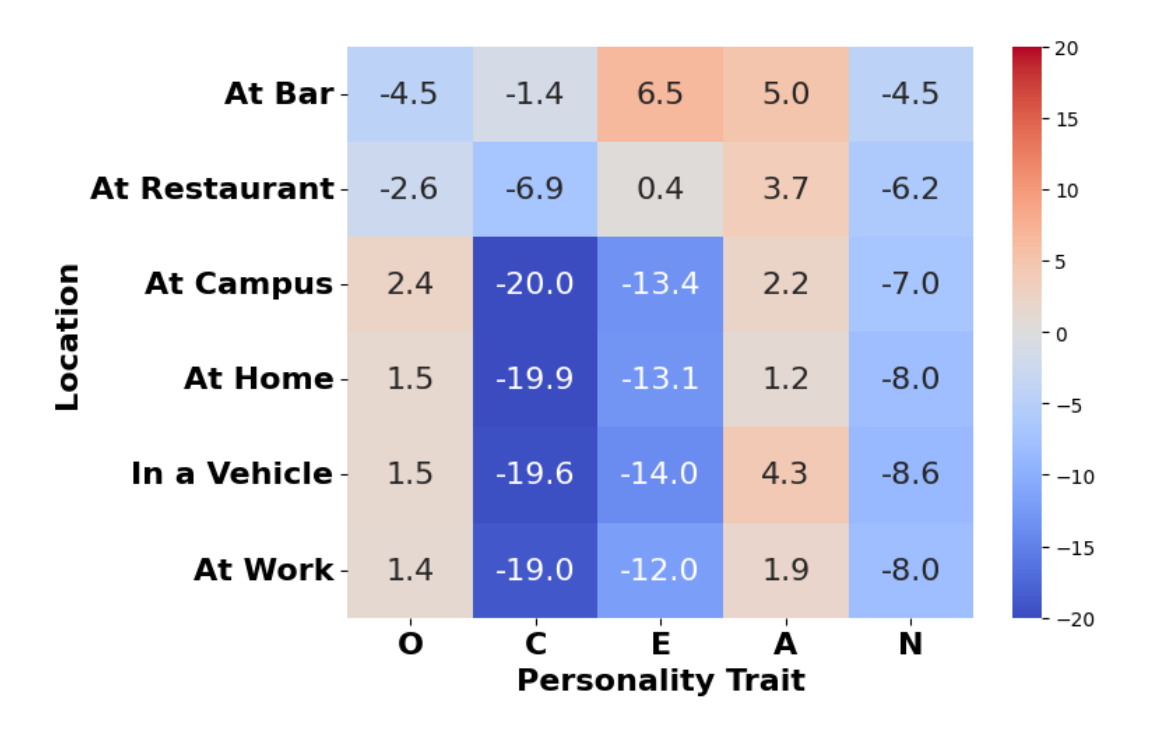}
        \caption{Personality change across locations for AutoGen.}
        \label{fig:AutoGen-location}
    \end{subfigure}
    \hfill
    \begin{subfigure}{0.45\linewidth}
        \centering
        \includegraphics[width=\linewidth]{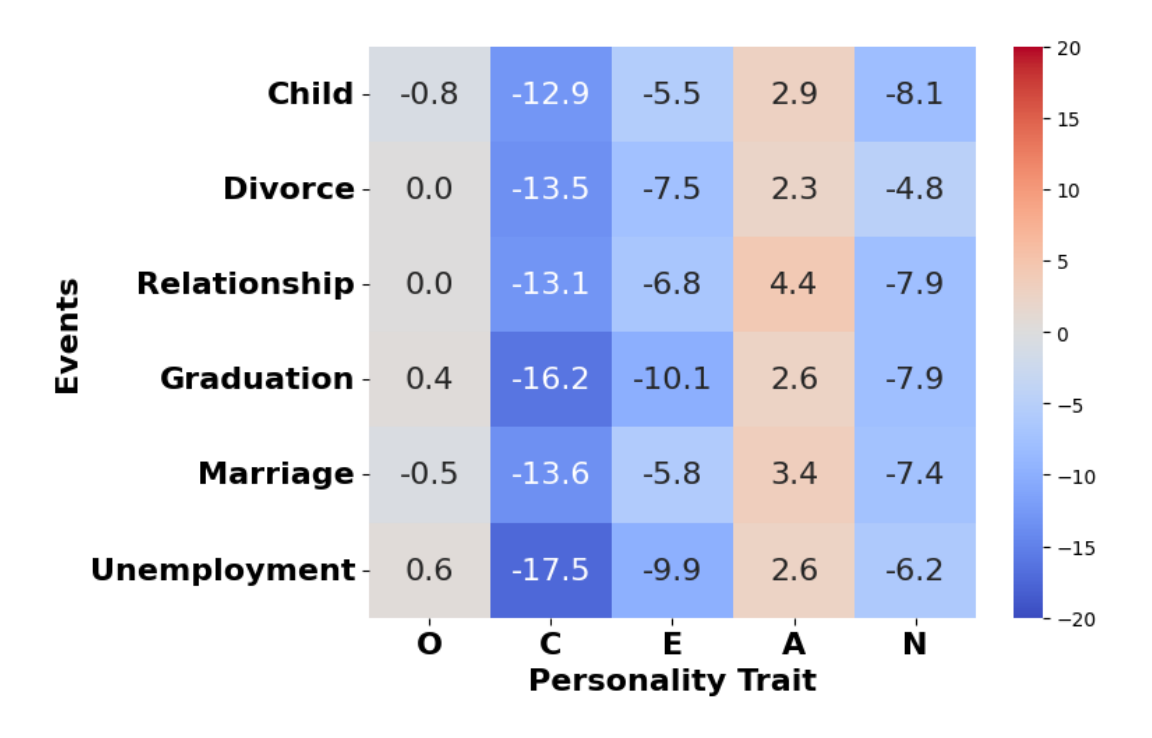}
        \caption{Personality change across events for AutoGen.}
        \label{fig:AutoGen-event}
    \end{subfigure}

    \caption{Comparison of average personality trait changes in AutoGen.}
    \label{fig:AutoGen-comparison}
\end{figure*}


This section provides a detailed breakdown of context-induced personality trait changes for all evaluated foundation models and agentic systems, complementing the summarized analyses in the main text.
\autoref{fig:gemini-comparison} to \autoref{fig:AutoGen-comparison} visualize the average trait change coefficients across both location-based and event-driven contexts.

Specifically, we observe relatively structured and bounded personality modulation.
For \textbf{Gemini 2.0 Flash} (\autoref{fig:gemini-comparison}), location-based contexts consistently induce positive shifts in Openness and Extraversion across all settings, with the strongest increases occurring in mobile and socially dynamic environments (e.g., `Vehicle', `Campus', up to +10). Event-driven contexts further amplify this pattern: relational events such as `Relationship' and `Marriage' lead to simultaneous increases in Extraversion and Agreeableness, while Neuroticism remains largely controlled except under `Unemployment'. These results indicate a stable but context-sensitive personality profile.

\textbf{Claude Sonnet 4} (\autoref{fig:Claude-comparison}) exhibits more heterogeneous responses. Location contexts trigger moderate increases in Openness and Extraversion in social settings (e.g., `At Bar'), but decreases in Conscientiousness appear in domestic contexts. Event-driven changes are substantially more pronounced: negative life events such as `Divorce' and `Unemployment' induce large drops in Conscientiousness and Agreeableness (often below -12), accompanied by sharp increases in Neuroticism (above +14). Compared with Gemini, Claude shows greater sensitivity to event polarity.

\textbf{GPT-OSS-120B} (\autoref{fig:GPTOSS-comparison}) demonstrates relatively moderate and asymmetric changes. Location contexts induce small positive shifts in Openness but mild declines in Extraversion and Agreeableness in private settings. Event-driven contexts reveal strong negative reactions to `Divorce' and `Unemployment', with large reductions in Conscientiousness and Extraversion and substantial increases in Neuroticism (up to +17.9). These patterns align with the main-text observation that foundation models differ in both stability and responsiveness.

Agentic systems display markedly amplified and less regulated personality shifts.
For \textbf{CAMEL} (\autoref{fig:Camel-comparison}), location-based contexts consistently reduce Conscientiousness and Extraversion (typically -4 to -6), while Openness remains mildly positive. Event-driven contexts follow a similar trend, with negative events causing further reductions in task-oriented traits and increases in Neuroticism.

\textbf{AutoGen} (\autoref{fig:AutoGen-comparison}) exhibits the most extreme instability among all evaluated systems. Across nearly all locations, Conscientiousness and Extraversion sharply decline (often approaching -20), while Neuroticism increases substantially. Event-driven contexts exacerbate this effect: `Unemployment' and `Graduation' trigger severe collapses in Conscientiousness (-16 to -18) and sustained reductions in Extraversion, indicating difficulty in maintaining coherent task-oriented behavior under prolonged contextual pressure.

In contrast to the above results, human personality changes reported in prior psychological studies (\autoref{tab:location-change}, and~\autoref{tab:event-change}) remain modest and balanced across both locations and life events, typically below 0.8 for locations and below 0.11 for events. While LLMs-especially foundation models-often mirror the direction of human personality change (e.g., increased sociability in relational contexts, elevated Neuroticism under stress), the magnitude of change in LLMs is substantially larger. This discrepancy is particularly pronounced in agentic systems, suggesting that current architectures capture context sensitivity but lack effective mechanisms for regulating trait intensity.

\begin{table*}[t]
\centering
\caption{Detailed location-driven personality trait changes for different LLM systems and humans.  Each cell reports the estimated trait change coefficient ($B$) induced by a location, with its standard error in parentheses.}
\label{tab:location-change-big}
\resizebox{1\linewidth}{!}{
\begin{tabular}{c|c|c|c|c|c|c}
\hline
\rowcolor[HTML]{F2F0F0} 
\textbf{Model}                          & \textbf{Location}      & \textbf{O}                            & \textbf{C}     & \textbf{E}                             & \textbf{A}     & \textbf{N}                             \\ \hline
                                        & \textbf{At Bar}        & 0.462 (0.102)                         & 1.247 (0.112)  & \cellcolor[HTML]{EFFFEF}1.503 (0.128)  & 0.516 (0.104)  & \cellcolor[HTML]{FDECEB}-0.247 (0.093) \\
                                        & \textbf{At Restaurant} & 0.237 (0.036)                         & 0.903 (0.039)  & 1.278 (0.044)                          & 0.629 (0.036)  & -0.227 (0.032)                         \\
                                        & \textbf{At Campus}     & 0.408 (0.021)                         & 0.849 (0.022)  & 0.365 (0.025)                          & 0.332 (0.021)  & 0.077 (0.019)                          \\
                                        & \textbf{Home}          & 0.355 (0.073)                         & 0.636 (0.079)  & 0.882 (0.089)                          & 0.595 (0.074)  & -0.113 (0.066)                         \\
                                        & \textbf{Vehicle}       & -0.170 (0.046)                        & 0.598 (0.050)  & 0.533 (0.057)                          & 0.204 (0.046)  & 0.090 (0.042)                          \\
\multirow{-6}{*}{\textbf{Human}}        & \textbf{Work}          & -0.048 (0.061)                        & 1.117 (0.065)  & 1.070 (0.074)                          & 0.871 (0.061)  & -0.001 (0.055)                         \\ \hline
                                        & \textbf{At Bar}        & -0.034 (0.334)                        & -0.017 (0.552) & -0.037 (0.382)                         & -0.003 (0.398) & -0.032 (0.458)                         \\
                                        & \textbf{At Restaurant} & -0.036 (0.327)                        & -0.012 (0.538) & -0.115 (0.396)                         & -0.010 (0.389) & -0.010 (0.472)                         \\
                                        & \textbf{At Campus}     & -0.035 (0.331)                        & -0.008 (0.547) & -0.122 (0.401)                         & -0.004 (0.395) & 0.018 (0.461)                          \\
                                        & \textbf{At Home}       & -0.051 (0.338)                        & -0.038 (0.556) & \cellcolor[HTML]{FDECEB}-0.169 (0.385) & -0.013 (0.401) & \cellcolor[HTML]{EFFFEF}0.008 (0.469)  \\
                                        & \textbf{In a Vehicle}  & -0.049 (0.325)                        & -0.029 (0.541) & -0.166 (0.392)                         & -0.026 (0.387) & 0.013 (0.473)                          \\
\multirow{-6}{*}{\textbf{Gemini}}       & \textbf{At Work}       & -0.038 (0.330)                        & -0.040 (0.549) & -0.150 (0.398)                         & -0.001 (0.392) & 0.003 (0.459)                          \\ \hline
                                        & \textbf{At Bar}        & 0.001 (0.370)                         & -0.023 (0.449) & -0.044 (0.523)                         & -0.010 (0.466) & -0.001 (0.615)                         \\
                                        & \textbf{At Restaurant} & 0.006 (0.358)                         & -0.020 (0.462) & -0.043 (0.542)                         & -0.003 (0.451) & -0.015 (0.638)                         \\
                                        & \textbf{At Campus}     & 0.003 (0.366)                         & -0.025 (0.444) & -0.043 (0.531)                         & -0.006 (0.457) & -0.011 (0.621)                         \\
                                        & \textbf{At Home}       & \cellcolor[HTML]{EFFFEF}0.008 (0.374) & -0.026 (0.469) & \cellcolor[HTML]{FDECEB}-0.059 (0.548) & -0.008 (0.463) & -0.004 (0.633)                         \\
                                        & \textbf{In a Vehicle}  & 0.006 (0.359)                         & -0.021 (0.452) & -0.036 (0.524)                         & -0.005 (0.455) & -0.023 (0.641)                         \\
\multirow{-6}{*}{\textbf{GPT-4o-mini}}  & \textbf{At Work}       & 0.006 (0.363)                         & -0.025 (0.460) & -0.052 (0.540)                         & -0.007 (0.461) & -0.003 (0.619)                         \\ \hline
                                        & \textbf{At Bar}        & -0.040 (0.334)                        & -0.020 (0.552) & -0.059 (0.382)                         & -0.003 (0.398) & -0.048 (0.458)                         \\
                                        & \textbf{At Restaurant} & -0.043 (0.327)                        & -0.013 (0.538) & -0.184 (0.396)                         & -0.012 (0.389) & -0.015 (0.472)                         \\
                                        & \textbf{At Campus}     & -0.042 (0.331)                        & -0.009 (0.547) & -0.195 (0.401)                         & -0.005 (0.395) & \cellcolor[HTML]{EFFFEF}0.027 (0.461)  \\
                                        & \textbf{At Home}       & -0.061 (0.338)                        & -0.042 (0.556) & \cellcolor[HTML]{FDECEB}-0.270 (0.385) & -0.016 (0.401) & 0.012 (0.469)                          \\
                                        & \textbf{In a Vehicle}  & -0.059 (0.325)                        & -0.032 (0.541) & -0.266 (0.392)                         & -0.031 (0.387) & 0.020 (0.473)                          \\
\multirow{-6}{*}{\textbf{GPT-OSS-120b}} & \textbf{At Work}       & -0.046 (0.330)                        & -0.044 (0.549) & -0.240 (0.398)                         & -0.001 (0.392) & 0.004 (0.459)                          \\ \hline
                                        & \textbf{At Bar}        & -0.043 (0.332)                        & -0.020 (0.553) & -0.054 (0.383)                         & -0.004 (0.399) & -0.045 (0.462)                         \\
                                        & \textbf{At Restaurant} & -0.045 (0.327)                        & -0.014 (0.540) & -0.167 (0.395)                         & -0.012 (0.391) & -0.014 (0.471)                         \\
                                        & \textbf{At Campus}     & -0.044 (0.331)                        & -0.009 (0.546) & -0.177 (0.401)                         & -0.005 (0.396) & \cellcolor[HTML]{EFFFEF}0.025 (0.460)  \\
                                        & \textbf{At Home}       & -0.064 (0.339)                        & -0.044 (0.556) & \cellcolor[HTML]{FDECEB}-0.245 (0.386) & -0.016 (0.402) & 0.011 (0.468)                          \\
                                        & \textbf{In a Vehicle}  & -0.061 (0.326)                        & -0.033 (0.542) & -0.241 (0.392)                         & -0.031 (0.388) & 0.018 (0.474)                          \\
\multirow{-6}{*}{\textbf{Claude}}       & \textbf{At Work}       & -0.048 (0.330)                        & -0.046 (0.550) & -0.218 (0.397)                         & -0.001 (0.392) & 0.004 (0.458)                          \\ \hline
\end{tabular}
}
\end{table*}


\begin{sidewaystable*}[t]
\centering
\caption{Detailed event-driven personality trait changes for different LLM systems and humans. 
Each cell reports the standardized effect size ($dE$) of an event on a given trait, along with its 95\% confidence interval.}
\label{tab:event-change-big}
\resizebox{\textheight}{!}{
\begin{tabular}{c|c|ccc|ccc|ccc|ccc|ccc}
\hline
\rowcolor[HTML]{F2F0F0} 
\cellcolor[HTML]{F2F0F0}                                 & \cellcolor[HTML]{F2F0F0}                                 & \multicolumn{3}{c|}{\cellcolor[HTML]{F2F0F0}\textbf{Human}} & \multicolumn{3}{c|}{\cellcolor[HTML]{F2F0F0}\textbf{Gemini 2.0 Flash}} & \multicolumn{3}{c|}{\cellcolor[HTML]{F2F0F0}\textbf{GPT-4o-mini}} & \multicolumn{3}{c|}{\cellcolor[HTML]{F2F0F0}\textbf{Claude Sonnet 4}} & \multicolumn{3}{c}{\cellcolor[HTML]{F2F0F0}\textbf{GPT-OSS-120b}} \\ \cline{3-17} 
\multirow{-2}{*}{\cellcolor[HTML]{F2F0F0}\textbf{Event}} & \multirow{-2}{*}{\cellcolor[HTML]{F2F0F0}\textbf{Trait}} & \textbf{dE}     & \textbf{95\%CI\_L}    & \textbf{95\%CI\_U}    & \textbf{dE}        & \textbf{95\%CI\_L}        & \textbf{95\%CI\_U}        & \textbf{dE}       & \textbf{95\%CI\_L}      & \textbf{95\%CI\_U}      & \textbf{dE}        & \textbf{95\%CI\_L}        & \textbf{95\%CI\_U}       & \textbf{dE}       & \textbf{95\%CI\_L}      & \textbf{95\%CI\_U}      \\ \hline
                                                         & \textbf{O}                                               & -0.078          & -0.167              & 0.012               & -1.162             & -2.499                  & 0.174                   & 0.135             & -0.613                & 0.884                 & -0.101             & -0.217                  & 0.018                  & -0.117            & -0.250                & 0.018                 \\
                                                         & \textbf{C}                                               & 0.022           & -0.079              & 0.124               & -1.569             & -3.149                  & 0.012                   & 0.014             & -0.73                 & 0.758                 & 0.031              & -0.095                  & 0.157                  & 0.037             & -0.103                & 0.177                 \\
                                                         & \textbf{E}                                               & -0.098          & -0.156              & -0.04               & -2.436             & -4.609                  & -0.263                  & 0.501             & -0.305                & 1.307                 & -0.147             & -0.228                  & -0.066                 & -0.186            & -0.280                & -0.092                \\
                                                         & \textbf{A}                                               & -0.09           & -0.225              & 0.044               & 0.374              & -0.632                  & 1.38                    & 0.146             & -0.603                & 0.896                 & -0.117             & -0.286                  & 0.052                  & -0.135            & -0.320                & 0.05                  \\
\multirow{-5}{*}{\textbf{Child}}                         & \textbf{N}                                               & -0.038          & -0.151              & 0.074               & 1.781              & 0.063                   & 3.5                     & 0.702             & -0.16                 & 1.563                 & -0.057             & -0.182                  & 0.068                  & -0.068            & -0.208                & 0.072                 \\ \hline
                                                         & \textbf{O}                                               & -0.062          & -0.138              & 0.015               & -1.575             & -3.16                   & 0.009                   & -0.336            & -1.109                & 0.436                 & -0.081             & -0.182                  & 0.02                   & -0.093            & -0.210                & 0.024                 \\
                                                         & \textbf{C}                                               & 0.096           & 0.033               & 0.158               & -4.473             & -8.178                  & -0.767                  & -1.763            & -3.084                & -0.441                & 0.134              & 0.05                    & 0.218                  & 0.163             & 0.061                 & 0.265                 \\
                                                         & \textbf{E}                                               & -0.024          & -0.165              & 0.117               & -18.528            & -33.385                 & -3.671                  & -1.679            & -2.958                & -0.4                  & -0.036             & -0.198                  & 0.126                  & -0.046            & -0.230                & 0.138                 \\
                                                         & \textbf{A}                                               & 0.008           & -0.047              & 0.062               & -2.496             & -4.712                  & -0.28                   & -1.258            & -2.336                & -0.18                 & 0.01               & -0.060                  & 0.08                   & 0.012             & -0.072                & 0.096                 \\
\multirow{-5}{*}{\textbf{Divorce}}                       & \textbf{N}                                               & -0.065          & -0.26               & 0.13                & 4.477              & 0.768                   & 8.185                   & 2.274             & 0.68                  & 3.868                 & -0.098             & -0.312                  & 0.116                  & -0.117            & -0.360                & 0.126                 \\ \hline
                                                         & \textbf{O}                                               & -0.035          & -0.067              & 0.137               & -0.383             & -1.391                  & 0.625                   & 0.161             & -0.589                & 0.911                 & -0.046             & -0.087                  & 0.178                  & -0.053            & -0.102                & 0.206                 \\
                                                         & \textbf{C}                                               & 0.154           & 0.044               & 0.265               & 2.613              & 0.312                   & 4.913                   & -0.029            & -0.773                & 0.715                 & 0.216              & 0.061                   & 0.371                  & 0.262             & 0.079                 & 0.445                 \\
                                                         & \textbf{E}                                               & 0.009           & -0.088              & 0.071               & 3.465              & 0.531                   & 6.4                     & 0.693             & -0.166                & 1.552                 & 0.014              & -0.110                  & 0.089                  & 0.017             & -0.127                & 0.161                 \\
                                                         & \textbf{A}                                               & -0.084          & -0.02               & 0.188               & 1.202              & -0.157                  & 2.562                   & 0.24              & -0.519                & 0.998                 & -0.109             & -0.026                  & 0.244                  & -0.126            & -0.030                & 0.282                 \\
\multirow{-5}{*}{\textbf{Relationship}}                  & \textbf{N}                                               & -0.116          & -0.316              & 0.085               & -4.245             & -7.775                  & -0.715                  & 0.215             & -0.541                & 0.971                 & -0.174             & -0.389                  & 0.041                  & -0.209            & -0.452                & 0.034                 \\ \hline
                                                         & \textbf{O}                                               & -0.175          & -0.316              & -0.033              & -0.542             & -1.596                  & 0.511                   & -0.06             & -0.804                & 0.685                 & -0.228             & -0.411                  & -0.045                 & -0.263            & -0.474                & -0.052                \\
                                                         & \textbf{C}                                               & -0.011          & -0.183              & 0.161               & 2.34               & 0.236                   & 4.444                   & 0.181             & -0.571                & 0.933                 & -0.015             & -0.220                  & 0.19                   & -0.019            & -0.244                & 0.206                 \\
                                                         & \textbf{E}                                               & -0.091          & -0.191              & 0.009               & 1.706              & 0.037                   & 3.376                   & 0.701             & -0.16                 & 1.562                 & -0.137             & -0.273                  & 0.009                  & -0.173            & -0.331                & -0.015                \\
                                                         & \textbf{A}                                               & -0.121          & -0.353              & 0.111               & 0.481              & -0.554                  & 1.515                   & 0.324             & -0.446                & 1.095                 & -0.157             & -0.459                  & 0.145                  & -0.182            & -0.520                & 0.156                 \\
\multirow{-5}{*}{\textbf{Marriage}}                      & \textbf{N}                                               & -0.023          & -0.109              & 0.063               & -2.476             & -4.678                  & -0.274                  & 0.136             & -0.613                & 0.884                 & -0.035             & -0.141                  & 0.071                  & -0.041            & -0.169                & 0.087                 \\ \hline
                                                         & \textbf{O}                                               & -0.041          & -0.285              & 0.202               & -0.523             & -1.571                  & 0.524                   & 0.079             & -0.666                & 0.825                 & -0.053             & -0.342                  & 0.236                  & -0.062            & -0.395                & 0.271                 \\
                                                         & \textbf{C}                                               & 0.128           & -0.142              & 0.398               & 0.854              & -0.324                  & 2.033                   & 0.178             & -0.574                & 0.93                  & 0.179              & -0.198                  & 0.556                  & 0.218             & -0.241                & 0.677                 \\
                                                         & \textbf{E}                                               & -0.011          & -0.079              & 0.057               & 0.817              & -0.345                  & 1.979                   & 0.707             & -0.156                & 1.571                 & -0.017             & -0.095                  & 0.061                  & -0.021            & -0.115                & 0.073                 \\
                                                         & \textbf{A}                                               & 0.07            & -0.15               & 0.289               & 0.212              & -0.763                  & 1.187                   & 0.215             & -0.541                & 0.97                  & 0.091              & -0.195                  & 0.377                  & 0.105             & -0.230                & 0.44                  \\
\multirow{-5}{*}{\textbf{Graduation}}                    & \textbf{N}                                               & -0.164          & -0.31               & -0.019              & -2.115             & -4.061                  & -0.169                  & 0.103             & -0.643                & 0.85                  & -0.246             & -0.423                  & -0.069                 & -0.295            & -0.495                & -0.095                \\ \hline
                                                         & \textbf{O}                                               & -0.039          & -0.191              & 0.113               & -1.057             & -2.336                  & 0.223                   & -0.167            & -0.918                & 0.584                 & -0.051             & -0.243                  & 0.141                  & -0.059            & -0.287                & 0.169                 \\
                                                         & \textbf{C}                                               & -0.058          & -0.097              & -0.019              & -4.847             & -8.842                  & -0.851                  & -1.732            & -3.038                & -0.426                & -0.081             & -0.124                  & -0.038                 & -0.099            & -0.153                & -0.045                \\
                                                         & \textbf{E}                                               & 0.053           & -0.151              & 0.258               & -12.981            & -23.412                 & -2.55                   & -1.459            & -2.63                 & -0.288                & 0.08               & -0.227                  & 0.387                  & 0.101             & -0.287                & 0.489                 \\
                                                         & \textbf{A}                                               & 0.182           & -0.298              & 0.662               & -0.821             & -1.984                  & 0.343                   & -0.661            & -1.51                 & 0.188                 & 0.237              & -0.365                  & 0.839                  & 0.273             & -0.422                & 0.968                 \\
\multirow{-5}{*}{\textbf{Unemployment}}                  & \textbf{N}                                               & -0.095          & -0.143              & -0.046              & 2.526              & 0.288                   & 4.764                   & 1.833             & 0.475                 & 3.192                 & -0.143             & -0.216                  & -0.070                 & -0.171            & -0.258                & -0.084                \\ \hline
\end{tabular}
}
\end{sidewaystable*}

\subsubsection{Pre-set Personality for LLM Systems}

\begin{sidewaystable*}[t]
\centering
\caption{Context-induced personality trait changes of GPT-4o-mini under different pre-set personality levels. The reported values are the mean change ($\Delta$) and standard deviation of each personality trait relative to the baseline condition.}
\label{tab:rq2-full}
\resizebox{\textheight}{!}{
\begin{tabular}{cc|ccc|ccc|ccc|ccc|ccc}
\hline
\rowcolor[HTML]{F2F0F0} 
\multicolumn{2}{c|}{\cellcolor[HTML]{F2F0F0}\textbf{Personality}}                 & \multicolumn{3}{c|}{\cellcolor[HTML]{F2F0F0}\textbf{O}} & \multicolumn{3}{c|}{\cellcolor[HTML]{F2F0F0}\textbf{C}} & \multicolumn{3}{c|}{\cellcolor[HTML]{F2F0F0}\textbf{E}} & \multicolumn{3}{c|}{\cellcolor[HTML]{F2F0F0}\textbf{A}} & \multicolumn{3}{c}{\cellcolor[HTML]{F2F0F0}\textbf{N}} \\ \hline
\multicolumn{2}{c|}{\textbf{Level}}                       & \textbf{High}    & \textbf{Medium}   & \textbf{Low}     & \textbf{High}     & \textbf{Medium}   & \textbf{Low}    & \textbf{High}    & \textbf{Medium}   & \textbf{Low}     & \textbf{High}    & \textbf{Medium}   & \textbf{Low}     & \textbf{High}    & \textbf{Medium}   & \textbf{Low}    \\ \hline
\multicolumn{1}{c|}{\multirow{6}{*}{\textbf{Location}}} & \textbf{At Bar}        & -0.211 (0.405) & 0.191 (0.461)   & 0.055 (0.979)  & -1.757 (1.042) & -0.012 (0.689)  & 1.013 (0.852) & -0.936 (0.898) & -0.604 (1.061)  & -0.230 (0.351) & -0.233 (0.482) & -0.068 (0.405)  & -0.021 (0.304) & -2.767 (1.636) & 1.028 (0.817)   & 1.621 (0.720)  \\
\multicolumn{1}{c|}{}                                   & \textbf{At Restaurant} & -0.163 (0.265) & 0.285 (0.556)   & 0.038 (0.906)  & -1.591 (1.036) & -0.011 (0.653)  & 0.968 (0.908) & -0.872 (0.905) & -0.654 (1.024)  & -0.202 (0.351) & -0.118 (0.412) & -0.000 (0.325)  & 0.029 (0.346)  & -2.747 (1.590) & 0.767 (0.753)   & 1.160 (0.516)  \\
\multicolumn{1}{c|}{}                                   & \textbf{At Campus}     & -0.117 (0.205) & 0.171 (0.407)   & 0.027 (0.955)  & -1.627 (1.076) & -0.086 (0.684)  & 0.896 (0.951) & -0.805 (0.812) & -0.780 (0.846)  & -0.131 (0.253) & -0.146 (0.434) & -0.030 (0.404)  & -0.003 (0.191) & -2.501 (1.619) & 0.774 (0.734)   & 1.110 (0.407)  \\
\multicolumn{1}{c|}{}                                   & \textbf{At Home}       & -0.210 (0.307) & 0.243 (0.406)   & 0.157 (1.071)  & -1.719 (1.055) & -0.044 (0.656)  & 0.913 (0.907) & -1.134 (0.972) & -0.969 (1.048)  & -0.275 (0.421) & -0.197 (0.406) & -0.040 (0.410)  & -0.005 (0.237) & -3.067 (1.460) & 1.146 (0.828)   & 1.655 (0.762)  \\
\multicolumn{1}{c|}{}                                   & \textbf{In a Vehicle}  & -0.132 (0.253) & 0.226 (0.480)   & 0.045 (0.936)  & -1.589 (1.028) & -0.025 (0.683)  & 0.919 (0.874) & -0.690 (0.780) & -0.615 (0.972)  & -0.163 (0.303) & -0.139 (0.378) & -0.014 (0.361)  & 0.012 (0.273)  & -2.536 (1.650) & 0.582 (0.677)   & 0.810 (0.370)  \\
\multicolumn{1}{c|}{}                                   & \textbf{At Work}       & -0.218 (0.342) & 0.264 (0.500)   & 0.075 (1.014)  & -1.725 (1.030) & -0.049 (0.749)  & 0.960 (0.821) & -1.059 (0.949) & -0.867 (1.075)  & -0.195 (0.372) & -0.206 (0.449) & -0.012 (0.414)  & -0.005 (0.240) & -2.763 (1.502) & 1.136 (0.826)   & 1.388 (0.637)  \\ \hline \hline
\multicolumn{1}{c|}{\multirow{6}{*}{\textbf{Event}}}    & \textbf{Child}         & -0.147 (0.248) & 0.280 (0.613)   & 0.084 (1.007)  & -1.379 (1.001) & 0.126 (0.818)   & 1.132 (1.009) & -0.377 (0.619) & -0.230 (1.180)  & -0.106 (0.381) & 0.032 (0.392)  & 0.067 (0.430)   & 0.047 (0.254)  & -3.029 (1.513) & 0.254 (0.677)   & 0.196 (0.406)  \\
\multicolumn{1}{c|}{}                                   & \textbf{Divorce}       & -0.457 (0.712) & 0.178 (0.429)   & 0.130 (1.076)  & -2.853 (2.090) & -0.446 (0.734)  & 0.648 (0.867) & -3.276 (2.256) & -2.949 (1.772)  & -0.666 (0.803) & -0.738 (0.839) & -0.217 (0.511)  & -0.121 (0.455) & -2.654 (1.657) & 3.617 (2.209)   & 6.171 (1.895)  \\
\multicolumn{1}{c|}{}                                   & \textbf{Relationship}  & -0.045 (0.143) & 0.330 (0.735)   & 0.056 (0.966)  & -1.237 (0.869) & 0.147 (0.780)   & 1.145 (0.932) & -0.152 (0.429) & 0.314 (1.304)   & -0.004 (0.193) & -0.016 (0.429) & 0.027 (0.406)   & 0.030 (0.189)  & -3.161 (1.594) & -0.244 (0.668)  & -0.001 (0.163) \\
\multicolumn{1}{c|}{}                                   & \textbf{Marriage}      & -0.126 (0.288) & 0.213 (0.524)   & -0.003 (0.968) & -1.270 (1.010) & 0.145 (0.819)   & 1.059 (1.029) & -0.132 (0.405) & 0.289 (1.332)   & -0.027 (0.206) & 0.010 (0.387)  & 0.050 (0.380)   & 0.036 (0.217)  & -2.872 (1.558) & -0.245 (0.602)  & 0.002 (0.156)  \\
\multicolumn{1}{c|}{}                                   & \textbf{Graduation}    & -0.072 (0.188) & 0.115 (0.398)   & 0.002 (0.922)  & -1.217 (1.034) & 0.068 (0.739)   & 1.040 (1.068) & -0.099 (0.383) & 0.077 (0.918)   & -0.040 (0.248) & -0.019 (0.388) & 0.016 (0.361)   & 0.034 (0.208)  & -2.680 (1.667) & -0.188 (0.540)  & 0.014 (0.124)  \\
\multicolumn{1}{c|}{}                                   & \textbf{Unemployment}  & -0.204 (0.439) & 0.266 (0.435)   & 0.129 (0.878)  & -2.051 (1.518) & -0.269 (0.537)  & 0.644 (0.611) & -1.462 (1.686) & -1.990 (1.776)  & -0.352 (0.620) & -0.307 (0.521) & -0.107 (0.375)  & -0.019 (0.331) & -1.984 (1.695) & 2.239 (1.471)   & 1.364 (1.103)  \\ \hline
\end{tabular}
}
\end{sidewaystable*}

Here we provide additional analyses for \autoref{tab:preset-base} and \autoref{tab:rq2-full}, complementing the main-text discussion on how preset personalities shape both baseline trait expression and subsequent context-induced personality trait changes.

\autoref{tab:rq2-full} further reveals that preset personality levels strongly modulate both the magnitude and often the direction of contextual trait change. A consistent pattern emerges across most location contexts: high preset traits tend to constrain subsequent shifts, whereas low presets more often amplify change. For example, under all six location contexts, Conscientiousness exhibits systematic decreases for High presets (roughly -1.6 to -1.8) but increases for Low presets (about +0.9 to +1.0), suggesting that `high-C' priors resist upward adjustment while `low-C' priors are more susceptible to context-driven compensation. Similar inverse tendencies appear for Extraversion, where High presets generally decrease (e.g., -0.7 to -1.1 across locations) while Low presets remain closer to zero or slightly negative, indicating that already-high sociability is harder to further increase under these locational cues.

Event-driven contexts accentuate this moderation effect, especially for negative events. Under `Divorce', Extraversion drops sharply under High presets (-3.28) but moves in the opposite direction under Low presets (+0.67). Neuroticism shows an even stronger polarity: High presets decrease (-2.65), Medium presets increase moderately (+3.62), and Low presets increase substantially (+6.17). This graded divergence suggests that preset traits not only shift the baseline but also define a `response regime' that governs how the model reacts to stress-inducing contexts. High presets appear to dampen reactivity, whereas low presets allow larger context-driven drift.

\subsubsection{Personality Impact on LLM Performance}

\begin{table*}[t]
\centering
\caption{AGIEval accuracy of GPT-OSS-120b under context-induced personality trait changes. 
\scriptsize{(`Avg. Improv.' indicates the average accuracy improvement of the model across the seven categories due to the personality changes. Bold indicates the best result, italic and underlined indicate the second-best result.)}}
\label{tab:ra3-performance}
\resizebox{1\linewidth}{!}{
\begin{tabular}{cc|cc|c|ccccccc|c}
\hline
\rowcolor[HTML]{F2F0F0} 
\textbf{Tasks}                   & \textbf{Trait}               & \textbf{Location} & \textbf{Event} & \textbf{Trait Change} & \textbf{aqua-rat} & \textbf{logiqa-en}             & \textbf{lsat-ar}               & \textbf{lsat-lr}               & \textbf{lsat-rc}               & \textbf{sat-en}                & \textbf{sat-math} & \textbf{Avg. Improv.}       \\ \hline
\rowcolor[HTML]{FFFFF0} 
\multicolumn{2}{c|}{\cellcolor[HTML]{FFFFF0}\textbf{Default}}   & /                 & /              & /               & 0.8898            & 0.3748                         & 0.1783                         & 0.4059                         & 0.5204                         & 0.5922                         & 0.9591            & /                              \\ \hline
\multicolumn{1}{c|}{\textbf{1}}  &                              & Vehicle           & Relationship   & 20.15\%         & 0.9592            & 0.3000 & 0.2609                         & 0.4200                         & 0.5400                         & 0.6042                         & 0.9800            & 0.0205                         \\
\multicolumn{1}{c|}{\textbf{2}}  &                              & Work              & Relationship   & 17.83\%         & 0.9592            & 0.4000                         & 0.2174                         & 0.3400 & 0.4694 & 0.6000                         & 0.9800            & 0.0065                         \\
\multicolumn{1}{c|}{\uline{\textit{\textbf{3}}}}  &
\multirow{-3}{*}{\uline{\textit{\textbf{O}}}} &
Bar &
Relationship &
17.05\% &
\uline{\textit{0.9600}} &
\uline{\textit{0.4600}} &
\uline{\textit{0.3333}} &
\uline{\textit{0.3400}} &
\uline{\textit{0.5200}} &
\uline{\textit{0.5918}} &
\uline{\textit{0.9800}} &
\uline{\textit{0.0378}} \\ \hline
\multicolumn{1}{c|}{\textbf{4}}  &                              & Home              & Unemployment   & -30.99\%        & 0.9592            & 0.3800                         & 0.1739 & 0.4000 & 0.5306                         & 0.6327                         & 0.9800            & 0.0194                         \\
\multicolumn{1}{c|}{\textbf{5}}  &                              & Vehicle           & Divorce        & -30.28\%        & 0.9592            & 0.3000 & 0.1556 & 0.4200                         & 0.6000                         & 0.5800                         & 0.9800            & 0.0106                         \\
\multicolumn{1}{c|}{\textbf{6}}  & \multirow{-3}{*}{\textbf{C}} & Vehicle           & Unemployment   & -28.17\%        & 0.9592            & 0.4400                         & 0.1200 & 0.3800 & 0.5918                         & 0.6600                         & 0.9800            & 0.0301                         \\ \hline
\multicolumn{1}{c|}{\textbf{7}}  &                              & Vehicle           & Unemployment   & -37.41\%        & 0.9592            & 0.4400                         & 0.1200 & 0.3800 & 0.5918                         & 0.6600                         & 0.9800            & 0.0301                         \\
\multicolumn{1}{c|}{\textbf{8}}  &                              & Campus            & Unemployment   & -35.25\%        & 0.9592            & 0.3600 & 0.1778 & 0.4600                         & 0.5102 & 0.6122                         & 0.9800            & 0.0198                         \\
\multicolumn{1}{c|}{\textbf{9}}  & \multirow{-3}{*}{\textbf{E}} & Home              & Unemployment   & -35.25\%        & 0.9592            & 0.3800                         & 0.1739 & 0.4000 & 0.5306                         & 0.6327                         & 0.9800            & 0.0194                         \\ \hline
\multicolumn{1}{c|}{\textbf{10}} &                              & Vehicle           & Divorce        & -31.85\%        & 0.9592            & 0.3000 & 0.1556 & 0.4200                         & 0.6000                         & 0.5800 & 0.9800            & 0.0106                         \\
\multicolumn{1}{c|}{\textbf{11}} &
&
Home &
Divorce &
-28.89\% &
\textbf{0.9592} &
\textbf{0.4000} &
\textbf{0.2000} &
\textbf{0.4400} &
\textbf{0.5400} &
\textbf{0.7083} &
\textbf{0.9800} &
\textbf{0.0439} \\
\multicolumn{1}{c|}{\textbf{12}} & \multirow{-3}{*}{\textbf{A}} & Work              & Divorce        & -28.15\%        & 0.9592            & 0.3600 & 0.2000                         & 0.4800                         & 0.5306                         & 0.6327 & 0.9800            & 0.0317                         \\ \hline
\multicolumn{1}{c|}{\textbf{13}} &                              & Vehicle           & Divorce        & 58.16\%         & 0.9592            & 0.3000 & 0.1556 & 0.4200                         & 0.6000                         & 0.5800 & 0.9800            & 0.0106                         \\
\multicolumn{1}{c|}{\textbf{14}} &                              & Bar               & Divorce        & 56.12\%         & 0.9388            & 0.4082                         & 0.2500                         & 0.4200                         & 0.4375 & 0.6600                         & 0.9800            & 0.0248                         \\
\multicolumn{1}{c|}{\textbf{15}} & \multirow{-3}{*}{\textbf{N}} & Home              & Unemployment   & 56.12\%         & 0.9592            & 0.3800                         & 0.1739 & 0.4000 & 0.5306                         & 0.6327                         & 0.9800            & 0.0194                         \\ \hline
\end{tabular}
}
\end{table*}

We further provide a detailed analysis of \autoref{tab:ra3-performance}, complementing the discussion by elaborating on the experimental design and fine-grained patterns across tasks and personality dimensions.

Specifically, for each Big Five personality trait, we first identify the contextual scenarios (location–event pairs) that induce the largest positive or negative deviations from the baseline personality state of GPT-OSS-120b. Reasoning performance is then evaluated under these selected contexts using AGIEval, which reports task-level accuracy across diverse reasoning categories, including mathematical reasoning (AQuA-RAT), logical reasoning (LogiQA, LSAT-AR, LSAT-LR), reading comprehension (LSAT-RC), and standardized language and math problems (SAT-EN, SAT-MATH). The default row corresponds to the same model evaluated without any contextual personality manipulation, serving as a consistent reference point. Red-highlighted cells indicate accuracy drops relative to this default condition, while unmarked cells indicate improvements.

\textbf{Trait-Specific Performance Sensitivity Across Tasks.} \autoref{tab:ra3-performance} reveals notable task-level heterogeneity in how personality shifts affect reasoning. For Openness, improvements are relatively broad-based: accuracy gains appear across both symbolic (AQuA-RAT, SAT-MATH) and language-heavy tasks (LogiQA, LSAT-RC), suggesting that higher Openness enhances general cognitive flexibility rather than benefiting a single task type. In contrast, declines in Conscientiousness disproportionately affect structured analytical tasks, such as LSAT-AR and LSAT-LR, where several accuracy values fall well below the default even when other tasks remain stable. This pattern aligns with the interpretation that task persistence and rule-following are critical for multi-step logical reasoning.

\textbf{Asymmetric Effects of Affective Traits.} The table also illustrates that Extraversion and Agreeableness exhibit more asymmetric effects. Performance degradation under reduced Extraversion is more pronounced for analytically demanding tasks (e.g., LSAT-AR), whereas reading comprehension and SAT-style tasks sometimes remain close to baseline. Agreeableness shows a narrower impact range: moderate decreases often have limited effect, but larger drops consistently coincide with declines in LSAT-RC and SAT-EN, indicating sensitivity in language understanding and contextual interpretation. These asymmetric patterns suggest that not all personality traits influence reasoning uniformly, and their effects depend on task structure and linguistic demands.

\textbf{Neuroticism as a Global Risk Factor.} Increases in Neuroticism stand out as the most consistently harmful condition in the table. Rows associated with large Neuroticism increases show red cells across nearly all tasks, including those that are otherwise robust under other trait changes. This uniform degradation suggests that elevated emotional instability introduces a global form of noise that undermines both symbolic reasoning and language-based inference, rather than selectively affecting specific task types.
\end{document}